\documentclass[manuscript,screen,colorlinks,allcolors=orange]{acmart}

\usepackage{xcolor}

\usepackage{enumitem}

\usepackage{amsmath,amsthm}
\usepackage{mathtools}
\usepackage{tabularx,array,threeparttable}
\usepackage{multirow}

\usepackage{scalerel}

\graphicspath{./figures}
\DeclareGraphicsExtensions{.pdf,.jpeg,.png}

\newcommand{\argmin}{\operatornamewithlimits{arg\,min}}
\newcommand{\argmax}{\operatornamewithlimits{arg\,max}}
\DeclarePairedDelimiter\abs{\lvert}{\rvert}%
\DeclarePairedDelimiter\norm{\lVert}{\rVert}%
\DeclarePairedDelimiter\dotp{\langle}{\rangle}
\DeclareMathOperator{\diag}{diag}

\newcommand\mydots{\hbox to 1em{.\hss.\hss.}}

\newtheorem{theorem}{Theorem}

\usepackage{algorithm,algpseudocode}

\newcommand{\ww}{\boldsymbol{w}}

\setcopyright{acmcopyright}
\copyrightyear{2024}
\acmYear{2024}
\acmDOI{XXXXXXX.XXXXXXX}

\begin{document}

\title{Batch Active Learning of Reward Functions from Human Preferences}

\author{Erdem B\i y\i k}
\email{biyik@usc.edu}
\orcid{0000-0002-9516-3130}
\affiliation{%
  \institution{Thomas Lord Department of Computer Science, University of Southern California}
  \city{Los Angeles}
  \state{California}
  \country{USA}
}

\author{Nima Anari}
\email{anari@cs.stanford.edu}
\affiliation{%
  \institution{Department of Computer Science, Stanford University}
  \city{Stanford}
  \state{California}
  \country{USA}
}

\author{Dorsa Sadigh}
\email{dorsa@cs.stanford.edu}
\affiliation{%
  \institution{Department of Computer Science, Stanford University}
  \city{Stanford}
  \state{California}
  \country{USA}
}

\renewcommand{\shortauthors}{B\i y\i k et al.}

\begin{abstract}
  Data generation and labeling are often expensive in robot learning. Preference-based learning is a concept that enables reliable labeling by querying users with preference questions. Active querying methods are commonly employed in preference-based learning to generate more informative data at the expense of parallelization and computation time. In this paper, we develop a set of novel algorithms, \emph{batch active preference-based learning} methods, that enable efficient learning of reward functions using as few data samples as possible while still having short query generation times and also retaining parallelizability. We introduce a method based on determinantal point processes (DPP) for active batch generation and several heuristic-based alternatives. Finally, we present our experimental results for a variety of robotics tasks in simulation. Our results suggest that our batch active learning algorithm requires only a few queries that are computed in a short amount of time. We showcase one of our algorithms in a study to learn human users' preferences.
\end{abstract}



\maketitle

\section{Introduction}

Machine learning algorithms have been quite successful in the past decade. A significant part of this success can be associated to the availability of large amount of labeled data.
However, collecting and labeling data can be costly and time-consuming in many fields such as speech recognition \cite{varadarajan2009maximizing}, recommendation systems \cite{biyik2023preference}, dialog control \cite{sugiyama2012preference}, image recognition \cite{sener2017geometric}, as well as in robotics \cite{sadigh2017active, palan2019learning, akrour2012april, jain2015learning}. Lack of labeled data is a common problem in many of these machine learning applications; however, it is particularly difficult in robot learning. Humans cannot reliably assign a \emph{success value} (reward) to a given robot trajectory, i.e., it is not obvious what labels need to be assigned to a particular robot behavior. When assigning labels is difficult, we often fall back to collecting expert demonstrations from humans to learn the desired behavior; however, this is also not easy in robotics applications as human experts often have a difficult time demonstrating optimal trajectories on a robot with high degrees of freedom \cite{akgun2012keyframe,losey2020controlling}, when there is uncertainty in the system or environment \cite{kwon2020when}, or under their cognitive biases about how a robot should operate \cite{basu2017you}.

To address the problem of assigning meaningful labels to single trajectories or the challenge of providing desirable and optimal trajectories, we instead focus on using \emph{preference-based learning} methods that query the users for their preferences in the form of comparison questions between multiple trajectories synthesized by the robot itself \cite{de2009preference,lee2021b,casper2023open,christiano2017deep,rafailov2023direct,liu2023efficient,hejna23contrastive}. These methods enable us to learn a regression model by only using the preferences of the users as opposed to relying on expert demonstrations.
	
Furthermore to address the lack of data in these applications, we leverage \emph{active} preference-based learning techniques, where we query the human with the most informative pairwise comparison question to recover their preferences of how a robot should act \cite{sadigh2017active,wilde2020active,tucker2020preference}. These preferences are often modeled using a reward function. However, active preference-based learning of reward functions can in practice be extremely time-inefficient, as these methods often require modeling a belief over a continuous reward function space and sampling from this belief. In addition, the states and actions in every trajectory that is shown to the human naturally are drawn from a continuous space, which often amplifies the time-inefficiency of these methods.

We thus propose using methods that generate a \emph{batch} of comparison queries optimized at the same time as opposed to generating queries one after the other.
These batch methods not only improve time-efficiency, but also have other computational benefits. For example, they can help when fitting the learning model is expensive, e.g., as in Gaussian processes \cite{biyik2020active, li2020roial,biyik2023active}, as the model should be retrained only after all queries in the batch are responded, rather than after every single query. In addition, these methods are parallelizable, which is a desirable feature when the robot is learning from multiple humans who share the same (or similar) preferences about how the task should be done.

While larger batches amplify these advantages, they can hurt data-efficiency, because new queries become less optimized with respect to the queries made earlier (and so the learned model so far). Hence, there is a direct tradeoff between \emph{the required number of queries} and the \emph{time it takes to generate each query}. Besides, it is challenging to decide how an informative batch must be generated.  While a batch of random queries hurts data-efficiency, finding the optimal batch is computationally intractable because it requires an exhaustive search over all possible human responses to the queries in the batch.

Ideally, we would like to develop an algorithm that requires only a few number of comparison queries while generating each query efficiently. In this work, we propose a new set of algorithms --- \emph{batch active preference-based learning} methods --- that balance this tradeoff between the number of queries it requires to learn human preferences and the time it spends on generation of each comparison query.

To this end, we actively generate each batch based on the data collected so far. 
Therefore, in our framework, we select and query $k$ pairs of trajectories, to be compared by the user or users, at once. Since $k$ queries are generated at once, our framework is parallelizable for data collection as opposed to standard active learning methods that require data to come sequentially.

\begin{figure*}[t]
	\centering
	\includegraphics[width=\textwidth]{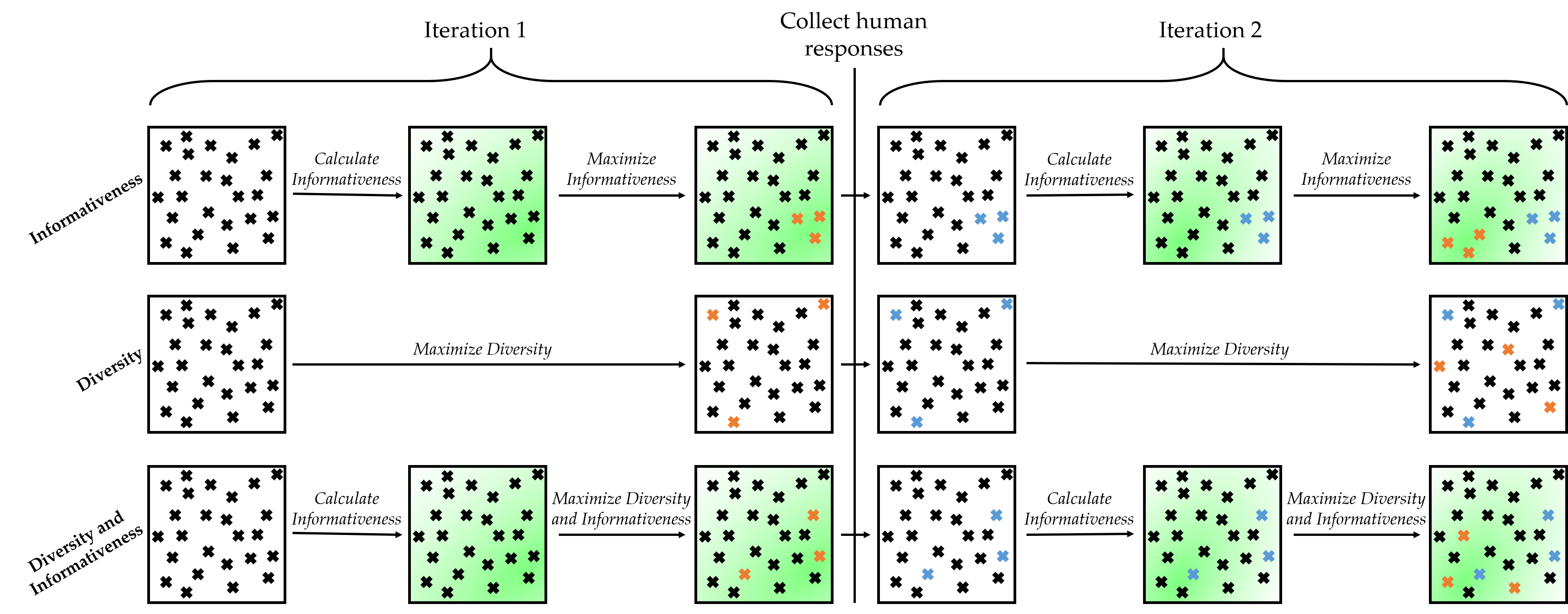}
	\caption{Batches should be both diverse and informative in batch active preference-based learning. Here, a hypothetical batch selection problem is visualized. Each cross represents a query. Similar queries are close to each other. Orange shows the queries selected in that iteration, and blue shows the queries for which the human responses have already been collected in the previous iterations. Green color represents informativeness: darker regions correspond to the queries with high informativeness based on the information collected until that iteration. \textbf{(Top)} Maximizing only informativeness generates batches that include very similar queries which, when queried together, carry redundant information. \textbf{(Middle)} Maximizing only diversity does not take informativeness into account at all, and so is wasteful as it selects some queries that are not informative. \textbf{(Bottom)} A good batch active learning algorithm should both select informative queries and avoid redundancy.}
	\label{fig:front_fig}
\end{figure*}

What makes batch active learning more difficult than standard active learning problems is that we cannot select the queries by simply maximizing their individual informativeness. Since a batch of queries is selected all at once, they must be selected without any information about the user responses to the queries within that batch. The batch active learning methods should then try to maximize the diversity between the queries in order to avoid selecting very similar queries in a single batch \cite{yang2015multi,cardoso2017ranked}. Therefore, a good batch active learning method must produce batches that consist of both dissimilar and informative queries. This is visualized in Figure~\ref{fig:front_fig}.

To this end, we summarize our contributions as\footnote{Note that parts of this work have been published at Conference on Robot Learning \cite{biyik2018batch} and as a preprint in \cite{biyik2019batch}. The DPP-based algorithm and all simulation experiment results are new compared to the conference paper.}:
\begin{enumerate}
\item Developing a batch active learning algorithm based on determinantal point processes (DPP) that leads to the highest performance by balancing the tradeoff between the informativeness and diversity of queries.
\item Designing a set of heuristic-based approximation algorithms for efficient batch active learning to learn about human preferences from comparison queries.
\item Experimenting and comparing approximation methods for batch active learning in complex preference based learning tasks.
\item Showcasing our framework in predicting human users' preferences in simulated autonomous driving and robotics tasks. 	
\end{enumerate}

For the rest of the paper, we will start with going over the related works in the literature and formalizing the problem. We will then present how standard active learning methods select queries. After introducing the general batch-mode active learning idea, we propose our methods for batch selection. First, we propose heuristic-based batch generation methods that avoid hyperparameter tuning. Next, we propose our primary method based on determinantal point processes. After proposing these different approaches, we present our experiments with both simulated and real users. We conclude the paper with a discussion of limitations and future work.
\section{Related Work}
\noindent\textbf{Inverse Reinforcement Learning.} There has been a lot of work on learning a model of the human preferences about the robots' trajectories through inverse reinforcement learning (IRL) \cite{ziebart2008maximum,levine2012continuous,sadigh2016planning,sadigh2017safe}. In these works, a reward function is learned directly from a human demonstrating how to operate a robot. However, learning a reward function from human demonstrations can be problematic for a few reasons. First, providing demonstrations for robots with higher degrees of freedom can be quite challenging even for human experts \cite{akgun2012keyframe,losey2020controlling}. Furthermore, human preferences tend to differ from their demonstrations \cite{basu2017you}. Therefore, prior work has extensively studied learning reward functions using other sources of information, e.g., human corrections \cite{bajcsy2018learning}, assessments \cite{shah2020interactive}, or rankings \cite{brown2020better,myers2022learning}. In this work, we learn the reward functions using \emph{preference queries} where the human is asked to simply compare two generated trajectories instead of demonstrating a trajectory.

\noindent\textbf{Batch Active Learning.} The problem of actively generating a batch of data is well-studied in other machine learning problems such as classification \cite{wei2015submodularity,elhamifar2016dissimilarity, yang2019single}, where decision boundaries may inform the active learning algorithms. While this may simplify the problem, it is not applicable in our setting, where we attempt to actively learn a reward function for dynamical systems using preference queries as opposed to data point -- label pairs where the labels are directly and persistently associated with the corresponding data points.

\noindent\textbf{Determinantal Point Processes.} While existing batch active learning methods are not readily applicable in our problem, we have the same challenge of generating both informative and diverse batches. For this, determinantal point processes (DPP) are a natural fit. DPPs are a mathematical tool that is often used for generating diverse batches from a set of items \cite{kulesza2012determinantal} and are used to generate batches in other machine learning applications, such as for improving the convergence of stochastic gradient descent \cite{zhang2017determinantal,zhang2019active}. Here, we propose using DPPs to generate not only diverse but also informative batches in active preference-based reward learning.

\noindent\textbf{Active Preference-based Learning.} Several works leveraged active preference-based techniques to synthesize pairwise comparison queries for the goal of efficiently learning humans' preferences \cite{akrour2011preference,furnkranz2012preference,sugiyama2012preference,wilson2012bayesian,biyik2022learning,ellis2024generalized}. However, there is a tradeoff between the time spent to generate a query at every time step and the number of queries required until converging to the human's preference reward function. 
Although actively synthesizing queries can reduce the total number of queries, generating each query can be quite time-consuming, which can make the approach impractical by creating a slow interaction with humans.

Most related to our work, \citet{sadigh2017active} and \citet{palan2019learning} have focused on learning reward functions by actively synthesizing comparison queries directly from the continuous space of states and actions, which caused these methods to be extremely slow. In fact, the participants of the user study by \citet{palan2019learning} raised concerns about the speed of the active preference-based reward learning framework.

Other methods have adopted the idea that the comparison queries can be actively selected from a pre-generated set of trajectories to reduce computational burden \cite{basu2019active}. While this improved the time-efficiency, the resulting method was still slow due to the fact that each and every query requires solving an optimization problem and the model has to be retrained after every human response. Moreover, none of these methods were parallelizable, i.e., they require the users to respond to the queries \emph{sequentially}, preventing parallel data collection. This negatively affects the use of these algorithms in the settings where multiple humans with shared preferences could provide responses to preference queries.

To this end, we propose a set of time-efficient \emph{batch} active learning methods, that balance between minimizing the number of queries and being time-efficient in its interaction with the human expert.
Batch active learning has two main benefits: i) generating a batch of queries can create a more \emph{time-efficient} interaction with the human, ii) the procedure can be \emph{parallelized} among multiple humans.
\section{Problem Statement}

\textbf{Modeling Choices.}
We start by modeling human preferences about how a robot should act. We model these preferences over the actions of a robot in a fully observable deterministic\footnote{We study deterministic dynamical systems to be able to compare with prior work \cite{sadigh2017active} without any major modifications. However, our methods can be easily extended to stochastic systems as long as the batch generation is performed over a predefined set of trajectories as we will elaborate in the text.} dynamical system $\mathcal{D}$. Let $f_{\mathcal{D}}$ denote the dynamics of the robot:
	\begin{align}
	x^{t+1} = f_{\mathcal{D}}(x^t, u^t)
	\label{eq:dynamics}
	\end{align}
Here $u^t$ denotes the action of the robot at time step $t$. The state $x^t$ evolves through the dynamics and the actions.

A finite trajectory $\xi \in \Xi$ is a sequence of state-action pairs $\left(\left(x^0, u^0\right)\dots \left(x^T, u^T\right)\right)$ over a finite horizon $t=0,1,\dots,T$. 
Here $\Xi$ is the set of feasible trajectories, i.e., trajectories that satisfy the dynamics of the system.

\begin{figure*}[t]
		\centering
		\includegraphics[width=0.8\textwidth]{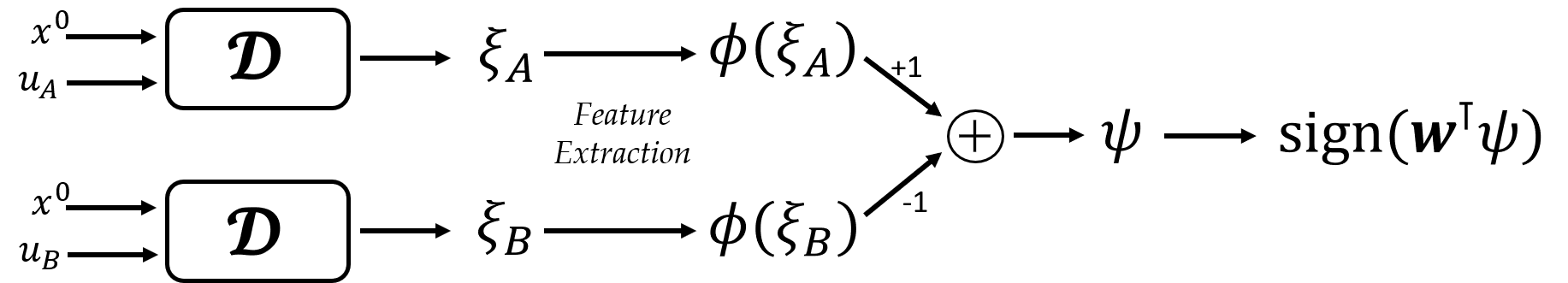}
		\caption{The schematic of the preferences based-learning problem starting from two sample inputs $(x^0, \mathbf{u}_A)$ and $(x^0, \mathbf{u}_B)$.}
		\label{fig:system}
	\end{figure*}

\textbf{Preference Reward Function.} We model human preferences through a preference reward function $R_H : \Xi \to \mathbb{R}$ that maps a feasible trajectory to a real number corresponding to a score for preference of the input trajectory. We assume the reward function is a linear combination of a given set of features over trajectories $\phi(\xi)\in\mathbb{R}^d$, where:
%
	\begin{align}
	R_H(\xi) := \ww^\top\phi(\xi)\:.
	\end{align}
	The goal of \emph{preference-based learning} is to learn $R_H(\xi)$, or equivalently the weights $\ww\in\mathbb{R}^d$ through preference queries from a human expert.
	A preference query is a question in the form of ``do you prefer trajectory $\xi_A$ or $\xi_B$?''. For any two trajectories $\xi_A$ and $\xi_B$, the human expert prefers $\xi_A$ over $\xi_B$ if and only if $R_H(\xi_A)>R_H(\xi_B)$.
	From this preference encoded as a strict inequality, we can conclude $\ww^\top\phi(\xi_A) > \ww^\top\phi(\xi_B)$ or equivalently: 
	\begin{equation}
		\ww^\top(\phi(\xi_A)-\phi(\xi_B))>0\:.
	\end{equation}
We use $\psi$ to refer to this difference: $\psi (\xi_A, \xi_B) := \phi(\xi_A)-\phi(\xi_B)$. Therefore, $\psi$ sufficiently characterizes the information we get from a query. Similarly, the sign of $\ww^\top\psi$ is sufficient to reveal the preference of the human expert for every trajectory pair $\xi_A$ and $\xi_B$. We thus let $I = \mathrm{sign}(\ww^\top\psi)$ denote the human's input (answer) to a query $\psi$. Figure~\ref{fig:system} summarizes the flow that leads to the human's preference $I$.

	In addition, the input from the human can be noisy due to the uncertainty of their preferences. A common noise model assumes human's preferences are probabilistic and can be modeled using a softmax function~\cite{sadigh2017active,christiano2017deep,holladay2016active}:
	\begin{align}
	P(I \mid \ww) &= \begin{cases}
	\frac{\exp(R_H(\xi_A))}{\exp(R_H(\xi_A)) + \exp(R_H(\xi_B))} & I=+1\\ 
	\frac{\exp(R_H(\xi_B))}{\exp(R_H(\xi_A)) + \exp(R_H(\xi_B))} & I=-1
	\end{cases}\nonumber\\
	&= \frac1{1+\exp(-I\ww^\top\psi)},
	\label{eq:human_noise}
	\end{align}
	where $I=\mathrm{sign}(\boldsymbol{w}^\top\psi)$ represents the preference of the human on the query $\psi$. This model led to successful inference of reward functions not only when preference data are provided by humans but also by vision-language models \cite{wang2024rl}.

\textbf{Approach Overview.} Our goal is to learn the human's reward function parameters $\ww$ in a both data-efficient and time-efficient way. To this end, we develop batch-active preference-based reward learning methods, that actively generate a batch of preference queries based on the previous queries and the human's responses to them.

In the next section, we first start with an overview of actively synthesizing queries where queries are optimized one by one for their informativeness. We then proceed with batch-active methods where we need to optimize both for informativeness (as in the non-batch setting) and diversity to avoid having similar (and so redundant) queries within a batch.
 
\section{Actively Synthesizing Pairwise Queries}
In active preference-based learning, the goal is to synthesize or search for the next pairwise comparison query to ask a human expert in order to maximize the information received. While optimal querying is NP-hard \cite{ailon2012active}, there exist techniques that pose the problem as a sequential optimization for which greedy solutions that work well in practice exist, e.g., volume removal \cite{sadigh2017active} and mutual information maximization \cite{biyik2019asking} methods.
Since \citet{biyik2019asking} identified failure cases of the former method and showed that maximizing mutual information yields consistently better results, we adopt this approach in our paper. We now describe this active preference-based reward learning approach.

	
The goal is to search for the human's preference reward function $R_H(\xi) = \ww^\top \phi (\xi)$ by actively querying the human. We  let $p(\ww)$ be the belief distribution of the unknown weight vector $\ww$. Since $\ww$ and $c\ww$ yield to the same true preferences for a positive constant $c$, we constrain the prior of the belief such that $\norm{\ww}_2\leq1$. Every query provides a human input $I$, which then enables us to perform a Bayesian update on this distribution:
	\begin{align}
	p(\ww\mid I) \propto p(I\mid\ww)p(\ww)\:,
	\label{eq:bayes_update}
	\end{align}
using the human preference model given in Eqn.~\eqref{eq:human_noise}. Since we do not know the shape of $p(\ww)$, we sample $M$ values from $p(\ww)$ using an adaptive Metropolis algorithm \cite{haario2001adaptive}. In order to speed up this sampling process, we approximate $p(I \mid \ww)$ with a log-concave function whose mode always evaluates to one:
	\begin{align}
	p(I \mid\ww) &= \min(1,\exp(I \ww^\top\psi))\:.
	\label{eq:human_noise_approx} 
	\end{align}
Based on \cite{biyik2019asking}, generating the next most informative query can be formulated as maximizing the mutual information between the human input $I$ and reward function parameters $\ww$ at every iteration. Put another way, we want to find the query that maximizes the difference between the prior entropy over $p(\ww)$ and the posterior entropy. We note that every query, i.e., a pair of trajectories $(\xi_A,\xi_B)$ is parameterized by the initial state of the system $x^0$, and the two sequences of actions $\mathbf{u}_A$ and $\mathbf{u}_B$ corresponding to $\xi_A$ and $\xi_B$, respectively, because the dynamics are deterministic with respect to Eqn.~\eqref{eq:dynamics}. Here, we assume the initial state of the system is the same between the two trajectories of a query (but not necessarily between queries) to make sure the trajectories are comparable to each other.
The query selection problem is then:
\begin{align}
\max_{x^0,\mathbf{u}_A,\mathbf{u}_B}\:H(\ww) - \mathbb{E}_I\left[H(\ww \mid I)\right]
\label{eq:optimization0}
\end{align}
with appropriate feasibility constraints to make sure the initial system state $x^0$ and the action sequences, $\mathbf{u}_A$ and $\mathbf{u}_B$, are feasible. Here, $H$ denotes information entropy \cite{cover1999elements}:
\begin{align}
H(\ww) = -\mathbb{E}_{\ww}\left[\log(p(\ww))\right]\:,
\end{align}
and we are trying to maximize the difference between the prior and the posterior entropies (mutual information) under the expected human input $I$.

Following the derivation for mutual information maximization in \cite{biyik2019asking}, we equivalently write the objective as:
\begin{align}
	\max_{x^0,\mathbf{u}_A,\mathbf{u}_B}\:\sum_{I\in\{-1,+1\}} \mathbb{E}_{\ww}\left[p(I\mid\ww)\log_2\left(\frac{p(I\mid\ww)}{\mathbb{E}_{\bar\ww}\left[p(I \mid \bar\ww)\right]}\right)\right]\:,
	\label{eq:optimization}
\end{align}
which we can approximately compute using the $\ww$ samples for the expectation terms. This query selection approach is similar to the expected value of information of the query \cite{krueger2016active,myers2023active} and the optimization can be solved using a Quasi-Newton method~\cite{andrew2007scalable}.

Actively generating queries significantly improves data-efficiency. However, it is not a time-efficient solution, since each and every query requires solving the optimization in \eqref{eq:optimization} and running the adaptive Metropolis algorithm \cite{haario2001adaptive} for sampling. Performing these operations for every single query might be quite slow and not very practical while interacting with a human expert in real-time. The human has to wait for the solution of optimization before being able to respond to the next query. Besides, all queries have to come sequentially, as each of them uses the information from all previous ones. This prevents collecting data in parallel where multiple people are available to provide preferences.

\subsection{Batch Active Learning}
Our insight is that we can in fact balance between the number of queries required for convergence to $R_H$ and the time required to generate each query. We construct this balance by introducing a \emph{batch active learning} approach, where $k$ queries are simultaneously generated at a time based on the current estimate of $\ww$. The batch approach can significantly reduce the total time required for the satisfactory estimation of $\ww$ at the expense of increasing the number of queries needed for convergence to true $R_H$.


To obtain a batch of queries that are informative, we need to find queries that have high mutual information values as computed by the objective of Eqn.~\eqref{eq:optimization}.  However, small perturbations of the inputs could lead to very minor changes in this objective value, and so continuous optimization of this objective can result in generating same or very similar queries within a batch. Besides, we want to increase the diversity of queries in a batch. We thus fall back to a discretization method. We discretize the space of queries by randomly sampling $K$ pairs of trajectories from the input space of $\xi = (x^0, \mathbf{u})$. While increasing $K$ may lead to more accurate optimization results, the computation time also increases linearly with $K$.
	

The batch active learning problem is then an optimization that attempts to find the $k$ queries out of $K$ that will maximize the mutual information between the human's responses and the reward function parameters $\ww$. Formally, we need to solve
\begin{align}
\max_{{\xi_1}_A,{\xi_1}_B,\dots,{\xi_k}_A,{\xi_k}_B \in \mathcal{K}}\:H(\ww) - \mathbb{E}_{I_1,I_2,\dots,I_k}\left[H(\ww \mid I_1,I_2,\dots,I_k)\right]
\label{eq:combinatorial_optimization0}
\end{align}
Although we can make a similar derivation for the objective as in \eqref{eq:optimization}, this is a combinatorial optimization problem that is often computationally hard (see \cite{cuong2013active} and \cite{chen2013near} for the proofs with similar objectives). Even though we first reduce the query set into a smaller set $\mathcal{X}$ of size $N$ by picking the queries which will individually maximize the mutual information (see Algorithm~\ref{alg:reduction} for the pseudo-code of this procedure), the solution still requires an exhaustive search which is intractable in practice as the search space is exponentially large \cite{guo2008discriminative}. For example, solving this combinatorial optimization with $k=10$ and $N=200$ would require evaluating the objective $C(200,10)\approxeq 2.25\times 10^{16}$ times.

\begin{algorithm}[ht]
	\caption{$\textsc{ReduceDataset}(\ww,\mathcal{K}, N)$}
	\label{alg:reduction}
	\begin{algorithmic}[1]
		\Statex \textbf{Input: } $\ww_1,\dots,\ww_M$\Comment{Sampled $\ww$ estimates}
		\Statex \textbf{Input: } $\mathcal{K}:=\left((\xi_{1_A},\xi_{1_B}),\dots,(\xi_{K_A},\xi_{K_B})\right)$\Comment{Dataset}
		\For{$i=1,\dots,K$}
		\State $\psi_i \gets \psi(\xi_{i_A},\xi_{i_B})$
		\State $q_i \gets H(\ww) - \mathbb{E}_I\left[H(\ww \mid I)\right]$
		\EndFor
		\State $\mathcal{X}\gets$ $\psi_i$'s with $N$ highest $q_i$ values \Comment{Reduction}
		\State $\mathbf{q} \gets q_i$ values corresponding to $\mathcal{X}$
		\State\Return $\mathcal{X}, \mathbf{q}$
	\end{algorithmic}
\end{algorithm}
	

\begin{algorithm}[h]
	\caption{Batch Active Preference-based Learning}
	\label{alg:batch_active_learning}
	\begin{algorithmic}[1]
		\State Generate query dataset $\mathcal{K}:=(\xi_{i_A},\xi_{i_B})_{i=1}^K$ w.r.t. \eqref{eq:dynamics}
		\For{$m = 1, 2, \dots$}
		\State Get $M$ samples $\ww\sim p(\ww)$
        \State $\mathcal{X}, \mathbf{q} \gets \textsc{ReduceDataset}(\ww,\mathcal{K}, N)$
		\State $A \gets$ a batch of size $k$ using $\ww$ from $\mathcal{X}$
		\State Get the human response for each query in $A$
		\State Update $p(\ww)$ according to \eqref{eq:bayes_update}
		\EndFor
		\State\Return $\mathbb{E}[\ww]$
	\end{algorithmic}
\end{algorithm}

In the subsequent sections, we present our batch generation algorithms that attempt to find approximately optimal batches without solving the combinatorial optimization and instead by using the individual mutual information values. We start with the most time-efficient heuristics with increasing complexity, and eventually present our premier method, DPP-based batch active learning, which leads to the best learning performance as we will present in our experiments. Algorithm~\ref{alg:batch_active_learning} gives an overview of the overall batch active preference-based learning approach: line 1 discretizes the space of queries, line 3 samples a set of $\ww$ from the belief distribution $p(\ww)$, and line 4 performs optional dataset reduction to work with the reduced set $\mathcal{X}$ for the current iteration instead of the full set $\mathcal{K}$. Line 5 produces a batch of queries, for which we present several methods in the subsequent sections. After the human responses are collected for the queries in the batch in line 6, a Bayesian update is performed to obtain the posterior belief distribution in line 7.
\section{Time-Efficient Batch Active Learning Methods}



We now describe a set of alternative methods in increasing order of complexity to provide approximations to the batch active learning problem. Figure~\ref{fig:method_visuals} visualizes each approach in this section for a small set of queries.

\subsection{Greedy Selection}
The simplest method to approximate the optimal batch generation is using a greedy strategy. In the greedy selection approach, we conveniently assume the $k$ different queries in a batch are independent from each other. Of course this is not a valid assumption, but the independence assumption allows us to choose the $k$-many maximizers of the objective of Eqn.~\eqref{eq:optimization} among the $K$ discrete queries.

\begin{figure*}[t]
	\centering
	\includegraphics[width=\textwidth]{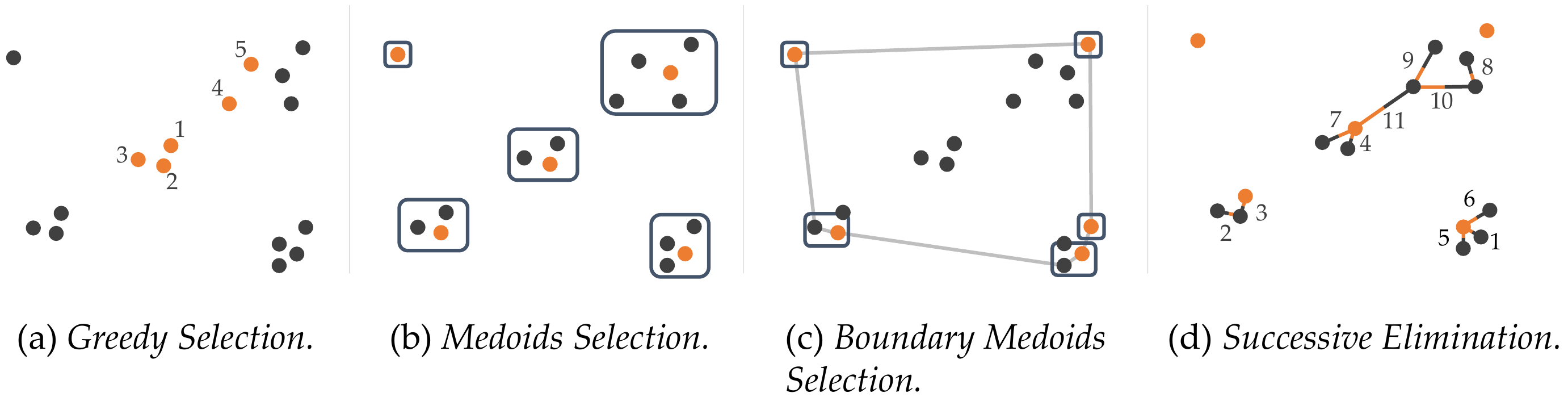}
	\caption{Visualizations of the batch generation process of the proposed time-efficient batch active learning algorithms. In each visual, a simple 2D space with 16 different $\psi$ values that correspond to the reduced set $\mathcal{X}$ is shown. The goal is to select a batch of $k=5$ that will near-optimally maximize the joint information gain. The selected queries are shown in orange. (a) Greedy Selection. (b) Medoids Selection. The points are selected based on the $k$-medoids clustering algorithm. (c) Boundary Medoids Selection. The clusters are chosen over the boundary of the convex hull of all samples. (d) Successive Elimination. One point is selected and another is eliminated based on pairwise comparisons of mutual information.
	}
	\label{fig:method_visuals}
\end{figure*}

While this method can easily be employed; it is suboptimal as similar or redundant queries can be selected together in the same batch because these similar queries are likely to lead to high mutual information values. For instance, as shown in Figure~\ref{fig:method_visuals}~(a), the $5$ orange queries chosen are all going to be very close to the center where mutual information values are high.

\subsection{Medoid Selection}
To avoid the redundancy in the batch created by the greedy selection, we need to increase the dissimilarity between the selected queries. We introduce an approach, \emph{Medoid Selection}, that leverages clustering as a similarity measure between the samples. In this approach, with the goal of picking the most dissimilar queries, we cluster $\psi$-vectors associated with the elements of the reduced set $\mathcal{X}$, whose elements are already invidual maximizers of mutual information, into $k$ clusters using standard Euclidean distance. We then restrict ourselves to only selecting one element from each cluster, which prevents us from selecting very similar trajectories.

One can think of using the well-known $k$-means algorithm \cite{lloyd1982least} for clustering and then selecting the centroid of each cluster. However, these centroids are not necessarily from the reduced set, so they can have lower mutual information values. More importantly, they might be infeasible, i.e., there might not be a pair of trajectories that produce the $\psi$ vectors corresponding to the centroids. 


Instead, we use the $k$-medoids algorithm \cite{kaufman1987clustering, bauckhage2015numpy} which again clusters the queries into $k$ sets. The main difference between $k$-means and $k$-medoids is that $k$-medoids enables us to select medoids as opposed to the centroids, which are queries \textit{in the set} $\mathcal{X}$ that minimize the average distance to the other queries in the same cluster.
While $k$-medoids is known to be a slower algorithm than $k$-means \cite{velmurugan2010computational}, efficient approximate algorithms exist \cite{bagaria2017medoids}. Figure~\ref{fig:method_visuals}~(b) shows the medoids selection approach, where $5$ orange queries are selected from the $5$ clusters.

\subsection{Boundary Medoid Selection}
We note that picking the medoid of each cluster is not the best option for increasing dissimilarity \textemdash instead, we can further exploit clustering to select queries more effectively. In the \emph{Boundary Medoid Selection} method, we propose restricting the selection to be only from the boundary of the convex hull of the reduced set $\mathcal{X}$.
If feasible, this selection criteria can separate out the selected queries from each other on average.
We note that when $d$, the dimension of $\psi$, is large enough compared to $k$, most of the clusters will have queries on the boundary. We thus propose the following modifications to the medoid selection algorithm.
The first step is to only select the queries that are on the boundary of the convex hull of the reduced set $\mathcal{X}$. We then apply $k$-medoids with $k$ clusters over the queries on the boundary and finally only accept the cluster medoids as the selected batch.
As shown in Figure~\ref{fig:method_visuals}~(c), we first find $k=5$ clusters over the points on the boundary of the convex hull of $\mathcal{X}$. We note that the number of queries on the boundary of convex hull of $\mathcal{X}$ can be larger than the number of queries needed in a batch, e.g., there are $7$ points on the boundary; however, we only select the medoids of the $5$ clusters created over these boundary queries shown in orange.

\subsection{Successive Elimination}
One of the main objectives of batch generation for active learning as described in the previous methods is to select $k$ queries that will maximize the average distance among them out of the $N$ queries in the reduced set $\mathcal{X}$. This problem is also referred to as \textit{max-sum diversification} in literature, which is known to be NP-hard \cite{gollapudi2009axiomatic, borodin2012max}. However, there exists a set of algorithms that provide approximate solutions~\cite{cevallos2017local}.

What makes our batch generation problem special and different from standard max-sum diversification is that we can compute the mutual information for each query. Mutual information is a metric that models how much we want a query to be in the final batch.
Thus, we propose a novel method that leverages the mutual information values to successively eliminate queries for the goal of obtaining a satisfactory diversified set. We refer to this algorithm as \emph{Successive Elimination}. At every iteration of the algorithm, we select two closest queries (in terms of Euclidean distance of their $\psi$ vectors) in the reduced set $\mathcal{X}$, and remove the one with lower mutual information value. We repeat this procedure until $k$ points are left in the set, resulting in the $k$ queries in our final batch, which efficiently increases the diversity among queries.

A pseudo-code of this method is given in Algorithm~\ref{alg:successive_elimination}. Figure~\ref{fig:method_visuals}~(d) shows the successive pairwise comparisons between two queries based on their corresponding mutual information. In every pairwise comparison, we eliminate one of the queries, shown with black edge, keeping the query connected with the orange edge. The numbers show the order of comparisons made before finding $k=5$ queries shown in orange.

\begin{algorithm}[ht]
	\caption{Successive Elimination}
	\label{alg:successive_elimination}
	\begin{algorithmic}[1]
		\State $\mathcal{X}, \mathbf{q} \gets \textsc{ReduceDataset}(\ww,\mathcal{K}, N)$
		\State $A \gets \mathcal{X}$ \Comment{Initialize the batch}
		\While{$\abs{A}>k$}
		\State $(\psi_i,\psi_j) \gets \argmin_{\psi_i,\psi_j\in A} \norm{\psi_i- \psi_j}_2$
		\If{$q_i < q_j$}
		\State Remove $\psi_i$ from $A$
		\Else
		\State Remove $\psi_j$ from $A$
		\EndIf
		\EndWhile
		\State\Return $A$
	\end{algorithmic}
\end{algorithm}

So far, we have presented four methods for batch selection in active preference-based reward learning: greedy, medoids, boundary medoids and successive elimination. While these methods are computationally efficient and easy-to-implement, they rely on rough heuristics. Determinantal point processes (DPP), on the other hand, provide a tractable mathematical procedure that we can elegantly adopt for batch selection. In the next section, we present our premier method based on DPPs.


\section{DPP-based Batch Active Learning}

Determinantal point processes (DPP) are a class of distributions that promote diversity. They are a natural fit for our problem as they can be tuned to balance the tradeoff between diversity and how desirable each item is. In our approach, we regard the set of queries as the item set of DPPs. We first start with presenting the necessary background on DPPs.

\subsection{Background}
A point process is a probability measure on a ground set $\mathcal{X}$ over finite subsets of $\mathcal{X}$. In our batch active preference-based learning framework, $\mathcal{X}$ is a set of queries. We let $\abs{\mathcal{X}}=K$.

An $L$-ensemble defines a DPP through a real, symmetric and positive semidefinite (PSD) $K$-by-$K$ kernel matrix $L$ \cite{borodin2005eynard}. Then, sampling a subset $X=A\subseteq \mathcal{X}$ has the probability
\begin{align}
P(X=A) \propto \det{L_A}
\label{eq:dpp}
\end{align}
where $L_A$ is an $\abs{A}$-by-$\abs{A}$ matrix that consists of the rows and columns of $L$ that correspond to the queries in $A$. For instance, if $A=\{i,j\}$, i.e. $A$ is a set consisting of $i^{\textrm{th}}$ and $j^{\textrm{th}}$ queries in $\mathcal{X}$, then
\begin{align*}
P(X=A)\propto L_{ii}L_{jj} - L_{ij}L_{ji}.
\end{align*}
We can consider $L_{ij}=L_{ji}$ as a similarity measure between the queries $i$ and $j$ in the set. The nonnegativeness of the second term in the above expression shows an example of \emph{repulsiveness} property of DPPs. This property makes DPPs the ubiquitous tractable point process to model negative correlations, and useful for generating diverse batches.

As $\det{L_A}$ can be positive for various $A$ with different cardinalities, we do not know $\abs{A}$ in advance. There is an extension of DPPs referred to as $k$-DPP where it is guaranteed that $\abs{A}=k$, and Eqn.~\eqref{eq:dpp} remains valid \cite{kulesza2011k}. In this work, we employ $k$-DPPs and refer to them as DPPs for the rest of the paper for brevity.



\begin{figure}[ht]
	\centering
	\includegraphics[width=0.5\textwidth]{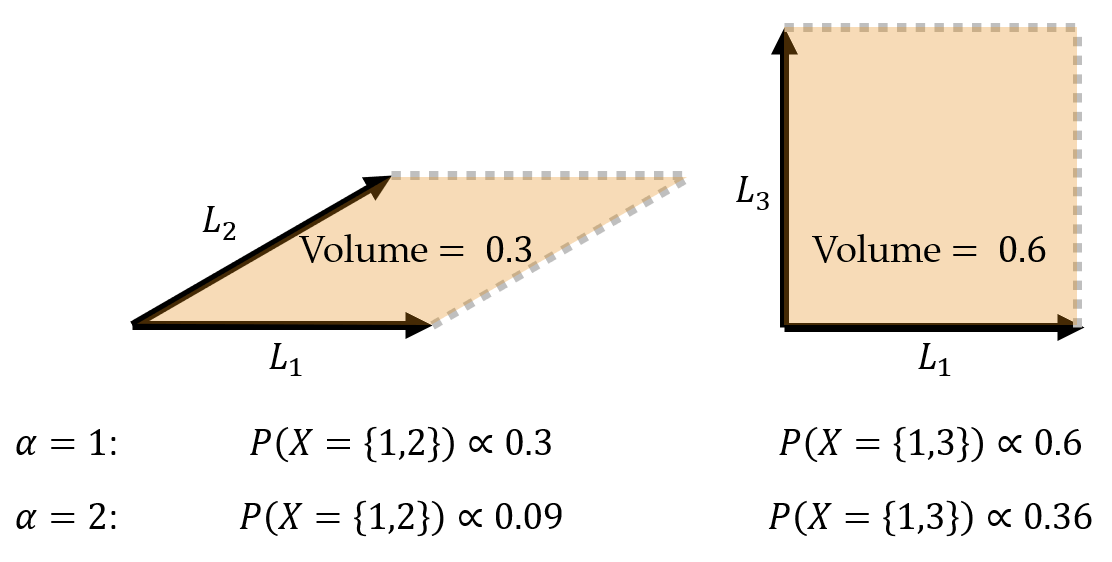}
	\caption{The effect of $\alpha$ is visualized. The columns of the matrix $L$ have the same length here; however $\{1,3\}$ is a more diverse set than $\{1,2\}$. When $\alpha=1$, $\{1,3\}$ is two times more likely to be sampled from the DPP distribution than $\{1,2\}$. When we increase $\alpha$ to $2$, this ratio increases to $4$, since more diverse sets are boosted against the less diverse sets.}
	\label{fig:dpp_alpha}
\end{figure}

Now, we explain what parameters we can have in an $L$-ensemble DPP. We note that
\begin{align*}
P(X=A)\propto \det L_A = \textrm{Vol}(\{L_i\}_{i\in A})\:,
\end{align*}
so the probability is proportional to the square of the associated volume.\footnote{Volume here refers to the volume of the parallelepiped spanned by the columns of $L$.} In fact, by using a generalized version of DPPs, we can approximately achieve \cite{anari2019log,mariet2018exponentiated}:
\begin{align}
P(X=A)\propto \textrm{Vol}^{\alpha}(\{L_i\}_{i\in A})\:,
\label{eq:alpha}
\end{align}
for $\alpha\geq0$. One can note that higher $\alpha$ enforces more diversity, because the probability of more diverse sets (larger volumes) will be boosted against the less diverse sets. We visualize this in Figure~\ref{fig:dpp_alpha}.

What remains is to construct the kernel matrix $L$. For this, we first define a matrix $S\in\mathbb{R}^{K\times K}$ whose entries measure the similarity between the queries. In our problem, each query has a feature difference vector $\psi$, and close $\psi$'s correspond to similar queries in terms of the information they provide. Therefore, we let
\begin{align}
    S_{ij} = \exp\left(-\frac{\norm{\psi_i-\psi_j}_2^2}{2\sigma^2}\right)\:,
\end{align}
where $\sigma$ is a hyperparameter. We then define the matrix $L$ as
\begin{align}
L_{ij} = q_i^{\gamma/\alpha}S_{ij}q_j^{\gamma/\alpha}\:,
\end{align}
which is guaranteed to be PSD by the construction of $S$. Here, $\gamma$ is another hyperparameter and $q_i\in\mathbb{R}_{\geq0}$ is the \emph{score} of $i^\textrm{th}$ query that represents how much we want that query in our batch. We use these scores to weight the queries based on their mutual information values, as computed by the objective of Eqn.~\eqref{eq:optimization}. By increasing $\gamma$ for fixed $\alpha$, we give more importance to the scores than diversity. This enables us set the tradeoff between informativeness and diversity.


\noindent\textbf{Relating the Mode of a DPP with High Diversity and Informativeness.} With proper tuning of $\alpha$ and $\gamma$, the batches that are both diverse and informative will have higher probabilities of being sampled. This motivates us to find the mode of the distribution, i.e., $\argmax_A \!P(X\!=\!A)$, which will guarantee informativeness and diversity. Another advantage of using the mode, instead of a random sample from the distribution, is the fact that it is significantly faster to approximate, even compared to the approximate sampling methods \cite{anari2016monte,li2016fast,mariet2018exponentiated,anari2019log}.

\subsection{Approximating the Mode of a DPP}

Finding the mode of a DPP exactly is NP-hard \cite{ko1995exact}. It is hard to even approximate it better than a factor of $2^{ck}$ for some $c\!>\!0$, under a cardinality constraint of size $k$ \cite{civril2013exponential}. Here, we discuss a greedy optimization algorithm to approximate the mode of a DPP.

In this approach, queries are greedily added to the batch. More formally, to approximate
\begin{align*}
\argmax_A P(X=A) = \argmax_A \textrm{Vol}^{\alpha}(\{L_i\}_{i\in A}),
\end{align*}
we greedily add queries to $A$. Let $A^{(l)}$ denote the set of selected queries at iteration $l$ of batch generation. We have
\begin{align*}
A^{(l+1)} = A^{(l)} \cup \{\argmax_j\textrm{Vol}^{\alpha}(\{L_i\}_{i\in A^{(l)}\cup\{j\}})\}\:,
\end{align*}
which we repeat until we obtain $k$ queries in $A$. \citet{ccivril2009selecting} showed that the greedy algorithm always finds a $k^{O(k)}$-approximation to the mode.

An important advantage of greedily approximating the mode is that the hyperparameter $\alpha$ becomes irrelevant, as it is just an exponent in the objective in every iteration of batch generation, unless trivially $\alpha=0$. This reduces the burden of hyperparameter tuning. 

While we use this greedy approach in our experiments for DPP-based batch generation, we also present a novel tractable algorithm, \emph{maximum coordinate rounding}, for approximating the mode of a DPP with better approximation ratio in Appendix~\ref{app:maximum_coordinate_rounding}. This algorithm is based on a convex relaxation of the optimization problem for finding the mode, which we solve using stochastic mirror descent. With the convex relaxation, we first perform the optimization as if we can take proportions of each query and then recursively select the queries by rounding. This algorithm achieves an $e^k$-approximation of the mode. The reason why we resort to the greedy approach in our experiments is because the maximum coordinate rounding algorithm, despite having polynomial time complexity, has much higher computational cost with large batch sizes in practice.

\subsection{Overall Algorithm}

Having presented the background in DPPs and the method to approximately find the DPP-mode, which corresponds to our diverse and informative batch, we are now ready to present our overall DPP-based batch active preference-based learning algorithm.

In this algorithm, we approximately compute the mode of the DPP distribution over the reduced set $\mathcal{X}$ as our batch. Algorithm~\ref{alg:dpp} presents the DPP-based method. The first for-loop (lines 2 through 7) constructs the DPP kernel, and the second part (lines 8 through 11) generates the batch by greedily approximating the mode of the constructed DPP. In our experiments, we set $\gamma=1$ and $\sigma$ to be the expected distance between two nearest points (in terms of Euclidean distance) when $k$ points are selected uniformly at random in the space $[0,1]^d$ where $d$ is the number of features, i.e., $d=\dim(\phi(\xi))$. This heuristic ensures a good kernel in line 4 of the algorithm as it is proportional to the Euclidean distance between queries.

\begin{algorithm}[ht]
	\caption{DPP-based Batch Generation}
	\label{alg:dpp}
	\begin{algorithmic}[1]
		\Require DPP hyperparameters $\sigma$, $\gamma$
		\State $\mathcal{X}, \mathbf{q} \gets \textsc{ReduceDataset}(\ww,\mathcal{K}, N)$
		\For{$\psi_i$ in $\mathcal{X}$}
		\For{$\psi_j$ in $\mathcal{X}$}
		\State $S_{ij} \gets \exp\left(-\frac{\norm{\psi_i-\psi_j}_2^2}{2\sigma^2}\right)$
		\State $L_{ij} \gets q_i^{\gamma}S_{ij}q_j^{\gamma}$
		\EndFor
		\EndFor
		\State $A \gets \emptyset$ \Comment{Initialize the batch}
		\For{$i=1,\dots,k$}
		\State $A \gets A \cup \{\argmax_j\det{L_{A\cup\{j\}}}\}$
		\EndFor
		\State\Return $A$
	\end{algorithmic}
\end{algorithm}

We make the code for all of our batch active learning methods available at \href{https://bit.ly/381brBK}{https://bit.ly/381brBK}. We also integrated them into APReL \cite{biyik2022aprel}, our comprehensive Python library for active preference based reward learning algorithms, which now enables experimenting batch active learning methods with various objectives, e.g., volume removal \cite{sadigh2017active}, max-regret \cite{wilde2020active}, etc. This is available at \href{https://github.com/Stanford-ILIAD/APReL}{https://github.com/Stanford-ILIAD/APReL}.

\section{Simulations and Experiments} \label{sec:experiments}

	\begin{figure*}
	\centering
	\includegraphics[width=0.8\textwidth]{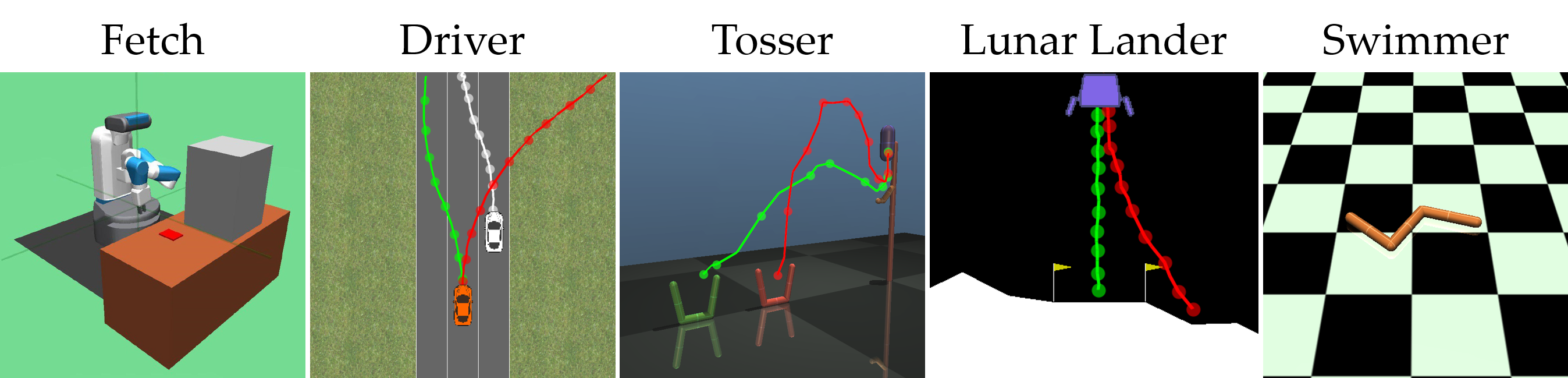}
	\caption{Simulation view of each environment. (a) Fetch, (b) Driver, (c) Tosser, (d) Lunar Lander, (e) Swimmer.}
	\label{fig:experiment_visuals}
\end{figure*}

\textbf{Experimental Setup.} We have performed several simulations and experiments to compare the methods we propose and to demonstrate their performance. Unless otherwise stated, we set batch size $k=10$, reduced query set size $N=200$ and number of $\ww$ samples $M=1000$ in all experiments.

\textbf{Alignment Metric.} For our simulations, we generate synthetic random $\ww_{\textrm{true}}$ vectors as our true preference vector. We have used the following alignment metric \cite{sadigh2017active} in order to compare non-batch active, batch active and random query selection methods, where all queries are selected randomly over all feasible trajectories.
\begin{align}
m = \frac{\ww_{\textrm{true}}^\top \hat{\ww}}{\norm{\ww_{\textrm{true}}}_2\norm{\hat{\ww}}_2}\:.
\end{align}
where $\hat{\ww}$ is $\mathbb{E}[\ww]$ based on the estimate of the learned distribution of $\ww$. We note that this alignment metric can be used to test convergence, because the value of $m$ being close to $1$ means the estimate of $\ww$ is very close to (aligned with) the true weight vector. 
In our experiments, we compare the methods using $m$ and the number of queries generated.

\textbf{Loglikelihood Metric.} Recent work has identified drawbacks of the alignment metric alogn with a discussion of other possible metrics \cite{wilde2022do}. Since our focus in this paper is reward learning, not reward optimization \cite{biyik2023active}, measuring the reward (or regret) of a policy that is optimized via the learned method is not a suitable metric. Instead, we use the loglikelihood metric which measures the loglikelihood of a held-out preference dataset \cite{biyik2020active,wilde2021learning,biyik2023active}.

\subsection{Tasks}
\begin{table}
	\centering
	\begin{threeparttable}
		\caption{Environment Properties}
		\label{tab:experiment_properties}
		\begin{tabular}{ c  c  c  c  c } 
			\hline
			\bfseries Task Name & $\boldsymbol{\dim(u^t)}$ & $\boldsymbol{T}$ & $\boldsymbol{d}$\\ 
			Fetch & 7 & 19 & 4 \\ 
			Driver & 2 & 5 & 4 \\
			Tosser & 2 & 2 & 4 \\
			Lunar Lander$^*$ & 2 & 5 & 6 \\
			Swimmer & 2 & 12 & 3 \\
			\hline
		\end{tabular}
		\begin{tablenotes}[para,flushleft]
			$^*$ Continuous version has been used.
		\end{tablenotes}
	\end{threeparttable}
\end{table}
We perform experiments in different simulation environments that are summarized in Table~\ref{tab:experiment_properties} with a list of the variables associated with every environment, where $d$ is the number of features, and $T$ is the horizon, i.e., the number of time steps. Note that these are all relatively short-horizon environments. This is because the optimization of queries in the non-batch active method, as it was described in \cite{sadigh2017active}, is over a $2\times(T\dim(u^t)) + \dim(x^0)$ dimensional space, where the factor $2$ is because we generate $2$ trajectories with the same initial state for each query. To keep the query synthesis tractable, we therefore modified the original environments to have relatively short horizons. Using a pregenerated dataset of queries as we do in the batch-active methods, however, eliminates this issue. Figure~\ref{fig:experiment_visuals} visualizes each of the experiment environments with some sample trajectories.

\textbf{Fetch.} Following \cite{palan2019learning}, we use the simulator for Fetch mobile manipulator robot \cite{wise2016fetch}, visualized in Figure~\ref{fig:experiment_visuals}~(a). For $\phi(\xi)$, we use features that correspond to average and final distances to the target object (red block), average distance to the table (brown block), and average distance to the obstacle (gray block).

\textbf{Driver.} We use the 2D driving simulator~\cite{sadigh2016planning}, shown in Figure~\ref{fig:experiment_visuals}~(b). We use features corresponding to distance to the closest lane, speed, heading angle, and distance to the other vehicle in the scenario. Two sample trajectories are shown in red and green in Figure~\ref{fig:experiment_visuals}~(b). In addition, the white line shows the fixed trajectory of the other vehicle on the road.

\textbf{Tosser.} We use MuJoCo's Tosser \cite{todorov2012mujoco} where a robot tosses a capsule-shaped object. The features we use are maximum horizontal range, maximum altitude, the sum of angular displacements at each timestep and final distance to closest basket of the object. The two red and green trajectories in Figure~\ref{fig:experiment_visuals}~(c) correspond to synthesized queries showing different preferences for what basket to toss the object to.

\textbf{Lunar Lander.} We use OpenAI Gym's Lunar Lander \cite{brockman2016openai} where a spacecraft is controlled. We use features corresponding to final heading angle, final distance to landing pad, total rotation, path length, final vertical speed, and flight duration. Two sample trajectories are shown in red and green in Figure~\ref{fig:experiment_visuals}~(d).

\textbf{Swimmer.} We use OpenAI Gym's Swimmer \cite{brockman2016openai}. We use features corresponding to horizontal displacement, vertical displacement, and total distance traveled. The environment is shown in Figure~\ref{fig:experiment_visuals}~(e).


\subsection{Comparison of Batch-Active Learning Methods}
We first quantitatively compare the batch-active methods we proposed with each other, as well as the combinatorial optimization problem posed in \eqref{eq:combinatorial_optimization0} for which we use simulated annealing \cite{bertsimas1993simulated,kochenderfer2019algorithms} as an approximate solution method. While simulated annealing could be run longer and longer to achieve better results, this would defeat the purpose of batch-mode active learning. Therefore, we limit its running time to match with the slowest algorithm among our proposed batch active methods. For each environment, we create a dataset of $K=500,\!000$ queries that consist of trajectories that are generated by taking random actions from a fixed initial state. Note that exploration within the environment is not an important issue here, because it is possible to learn the true reward function by only comparing suboptimal trajectories as long as the queries cover the feature space well, i.e., we do not need to ensure there are successful trajectories in the query dataset.

Independently for each environment, we randomly generated $100$ different reward functions ($\ww_{\textrm{true}}$ vectors) for tests of all methods. We then simulated noiseless users, who always reveal their true preferences in order to eliminate the effect of noise in the results. However, the learning methods still adopted the noisy user model we presented in Eqn.~\eqref{eq:human_noise_approx}.

\begin{figure}[ht]
	\centering
	\includegraphics[width=\textwidth]{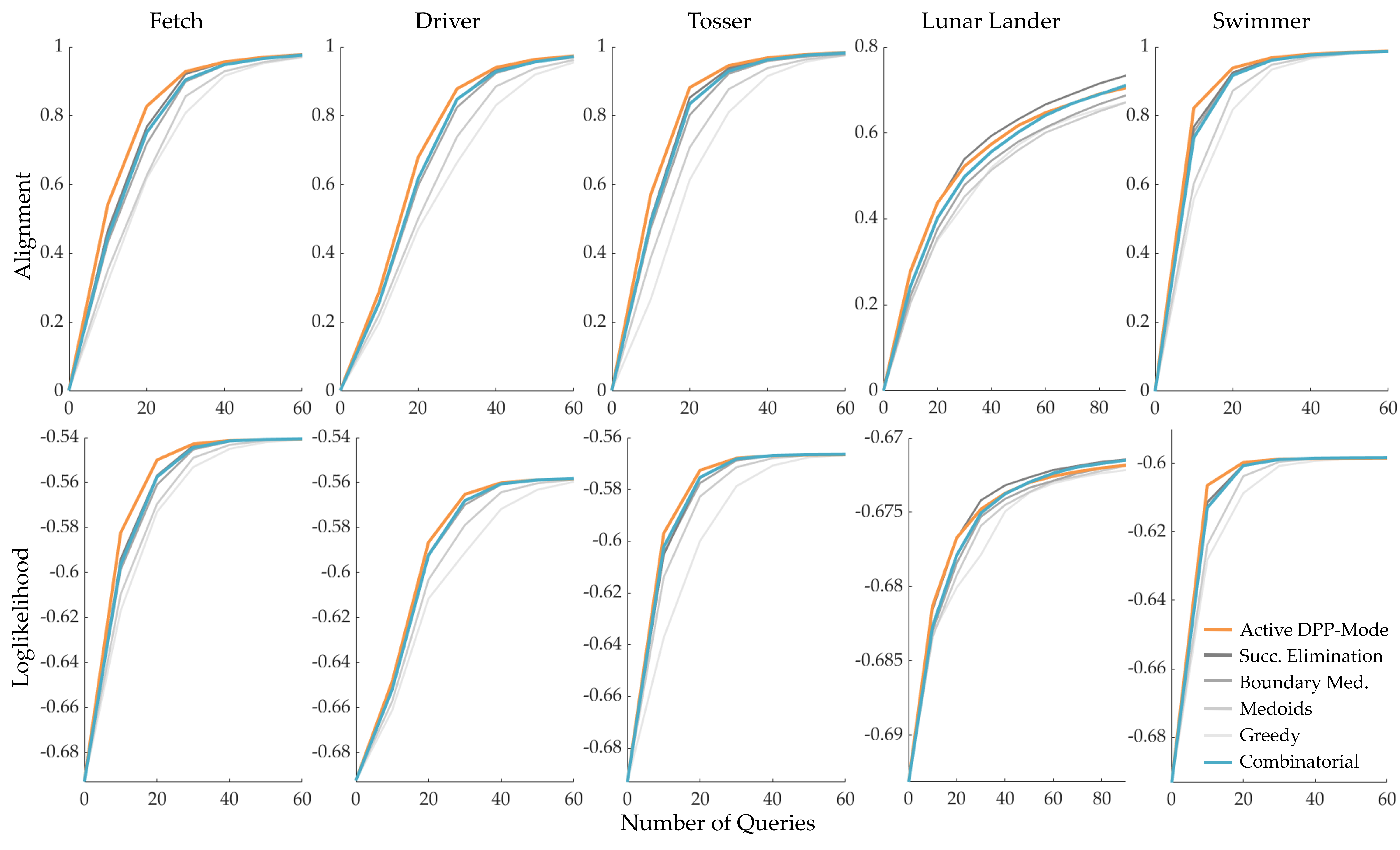}
	\caption{Batch-active learning methods are compared.}
	\label{fig:dpp_results}
\end{figure}
For each simulated reward function during our tests, we ran $6$ batch generations with each method, summing up to $60$ pairwise comparison queries. For Lunar Lander where learning is more dificult due to larger feature dimensionality, we ran $9$ batches ($90$ queries). We demonstrate the results in Figure~\ref{fig:dpp_results}. Since the reward functions are paired between the methods, we use Wilcoxon signed-rank tests \cite{wilcoxon1945individual} over the \emph{area under the curve} for the alignment metric as it was done in previous work \cite{myers2022learning}. Our results suggest that the DPP-based method significantly outperforms all other methods, including the combinatorial optimization via simulated annealing, in all environments ($p<0.005$) except Lunar Lander where successive elimination is the best-performing method.

Among the heuristic-based batch active learning methods we proposed, successive elimination method significantly outperforms the others ($p<0.005$ in all except Swimmer where it is only $p<0.05$ against the boundary medoids method). It also outperforms the combinatorial optimization in all environments ($p<0.005$) except Driver where both methods perform comparably.

Combinatorial optimization and boundary medoids both significantly outperform medoids and greedy methods in all environments ($p<0.005$). While boundary medoids method significantly outperforms combinatorial optimization in Swimmer ($p<0.05$), combinatorial optimization is significantly better in the other environments ($p<0.05$ in Fetch and $p<0.005$ in others). Finally, medoids method significantly outperforms the greedy method in all environments ($p<0.005$) except Lunar Lander where they perform comparably.

Overall, these results show us the ranking of batch active learning methods from the best to the worst are as follows:
\begin{enumerate}[nosep]
	\item Active DPP-Mode
	\item Successive Elimination
    \item Combinatorial Optimization via Simulated Annealing
	\item Boundary Medoids
	\item Medoids
	\item Greedy
\end{enumerate}

\subsection{Comparison to Non-Batch Active Learning}
We next investigated the average time it required to generate one query. For this, we again took a dataset of $K=500,\!000$ queries. We recorded the batch generation times, and divided it by $k=10$ to get the time per query. To show the advantage of batch-active learning methods, we also ran the same analysis on the non-batch active learning approach. We had to exclude the non-batch active method in the \textit{Fetch} environment, as it was not able to synthesize queries in reasonable amounts of time due to the large action space. The results are shown in Table~\ref{tab:average_query_times}. It can be seen that batch active learning methods lead to a great decrease in query generation times compared to the non-batch method.

\begin{table*}
	\centering
	\begin{threeparttable}
		\caption{Average Query Generation Times (seconds)}
		\label{tab:average_query_times}
		\begin{tabular}{ c | c | c  c  c  c  c c } 
			\hline
			\multirow{2}{*}{\bfseries Environment} & \multirow{2}{*}{\bfseries Non-Batch} & \multicolumn{6}{c}{\bfseries Batch Active Learning}\\
			& & \emph{Combinatorial} & \emph{Greedy} & \emph{Medoids} & \emph{Boundary Med.} & \emph{Succ. Elimination} & \emph{DPP}\\
			\hline
            \textit{Fetch} & N/A & 3.70 & 3.34 & 3.33 & 3.36 & 3.67 & 3.36 \\
			\textit{Driver} & 36.42 & 3.28 & 2.85 & 2.87 & 2.87 & 3.23 & 2.89 \\ 
			\textit{Tosser} & 46.21 & 3.31 & 2.88 & 2.89 & 2.90 & 3.21 & 2.89 \\
			\textit{LunarLander} & 69.42 & 3.33 & 3.03 & 3.02 & 3.03 & 3.36 & 3.04 \\
			\textit{Swimmer} & 146.32 & 3.29 & 2.87 & 2.89 & 2.89 & 3.22 & 2.89 \\
		\end{tabular}
	\end{threeparttable}
\end{table*}

As we observed that our DPP-based method generates highly informative queries in a time-efficient way, we now compare its performance to the non-batch active learning approaches. Specifically, we assessed its performance against non-batch active learning and random query selection where queries are selected uniformly at random from $\mathcal{K}$.

\begin{figure*}
	\centering
	\includegraphics[width=\textwidth]{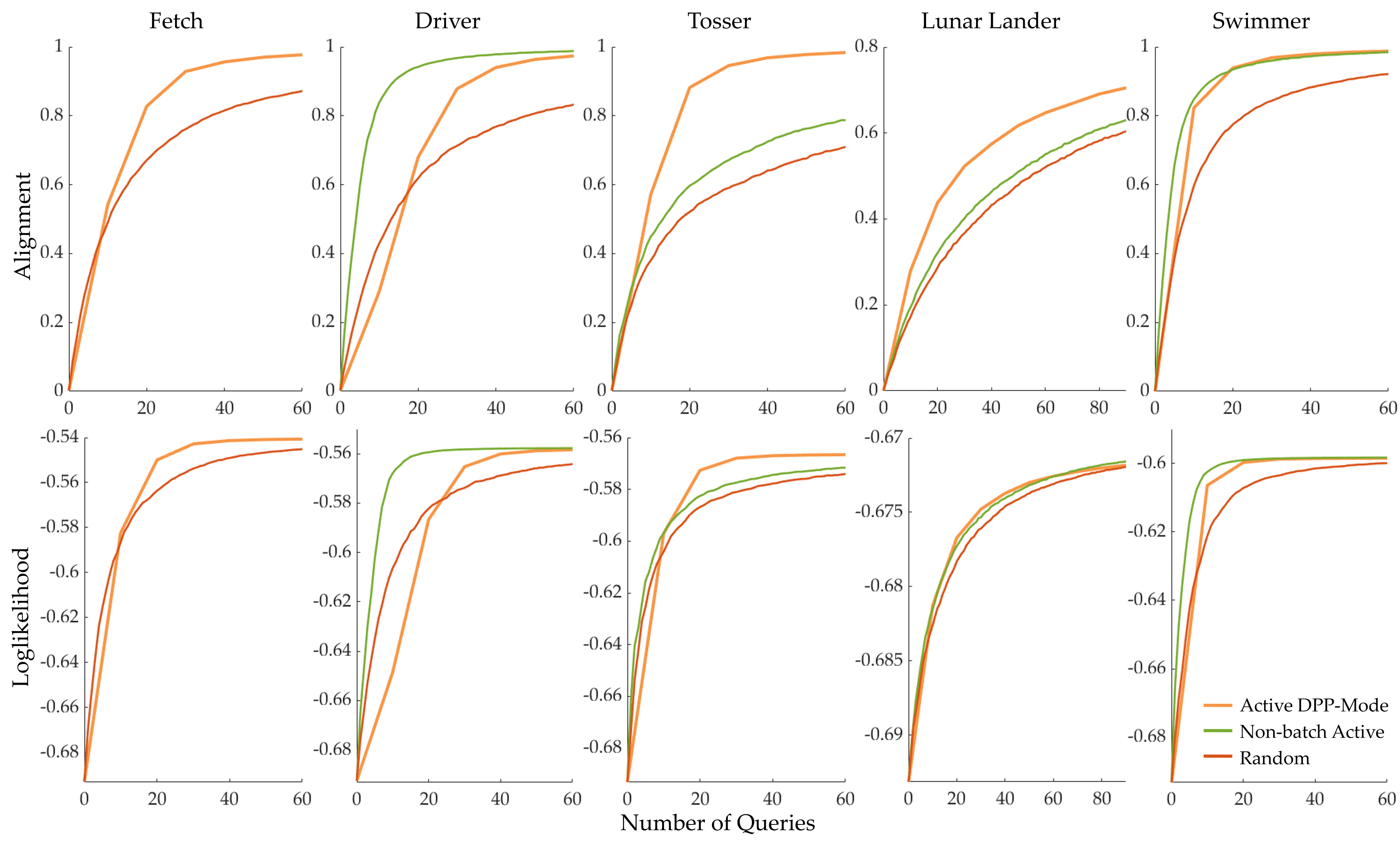}
	\caption{The performance the algorithms is shown. The non-batch active method performs poorly on Lunar Lander and Tosser.}
	\label{fig:query_count}
\end{figure*}

\begin{figure*}
	\centering
	\includegraphics[width=\textwidth]{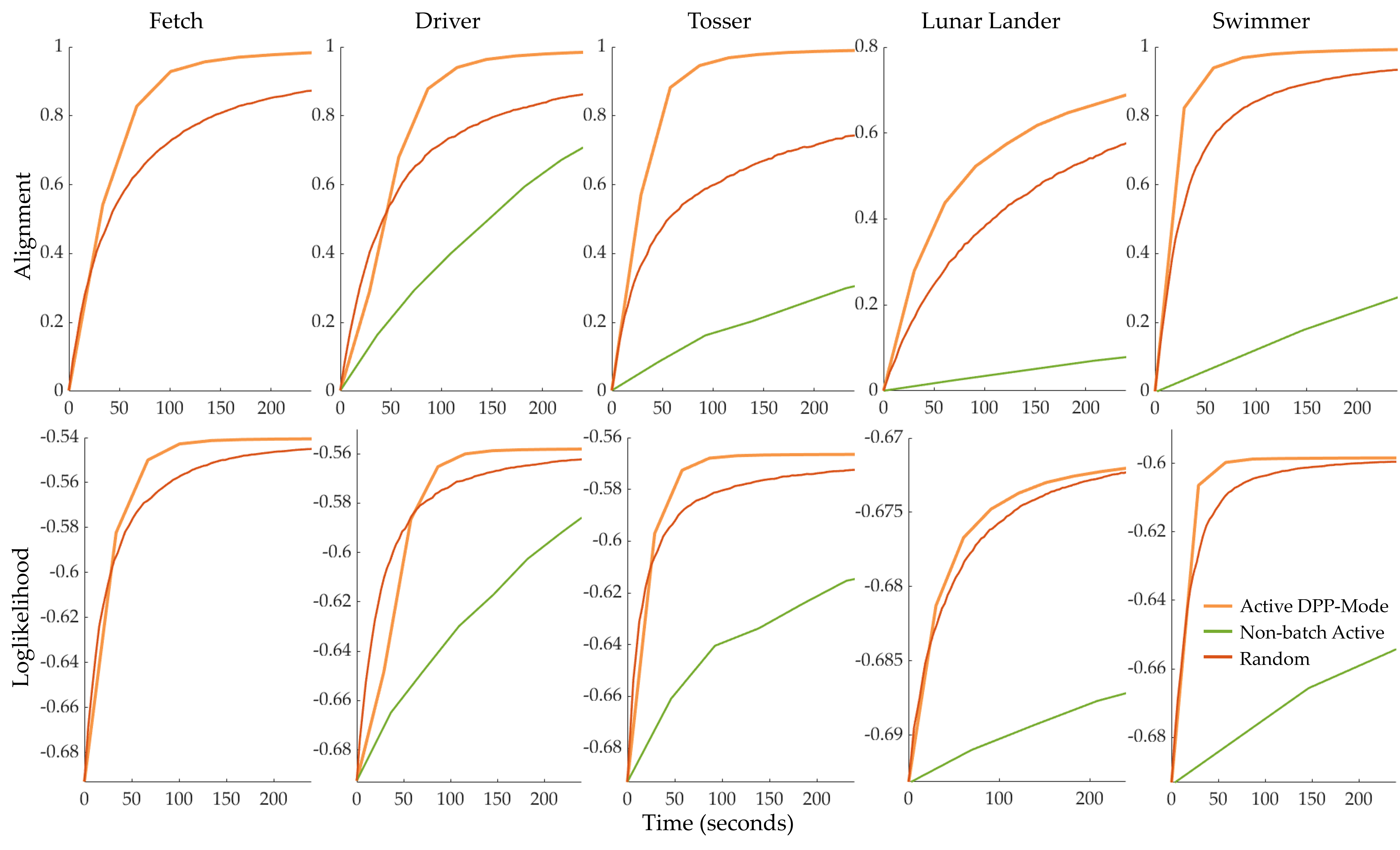}
	\caption{Alignment and loglikelihood as a function of time are plotted for each environment. Non-batch active learning method is slow due to the optimization and adaptive metropolis algorithm involved in each iteration, whereas random querying performs poorly due to redundant queries. The DPP-based method clearly outperforms both of them.}
	\label{fig:timing}
\end{figure*}

We show the results in Figures~\ref{fig:query_count} and \ref{fig:timing}. Figure~\ref{fig:query_count} shows the convergence to the true weights $\ww_{\textrm{true}}$ in terms of alignment and loglikelihood metrics as the number of queries increases. It is interesting to note that non-batch active learning performs suboptimally in \textit{LunarLander} and \textit{Tosser}. This is because the continuous query synthesis by optimizing over the initial state and the sequences of actions fails in these environments. Our DPP-based method, on the other hand, sidesteps this problem due to discrete optimization over the query set.

\begin{figure}[ht]
	\centering
	\includegraphics[width=0.9\textwidth]{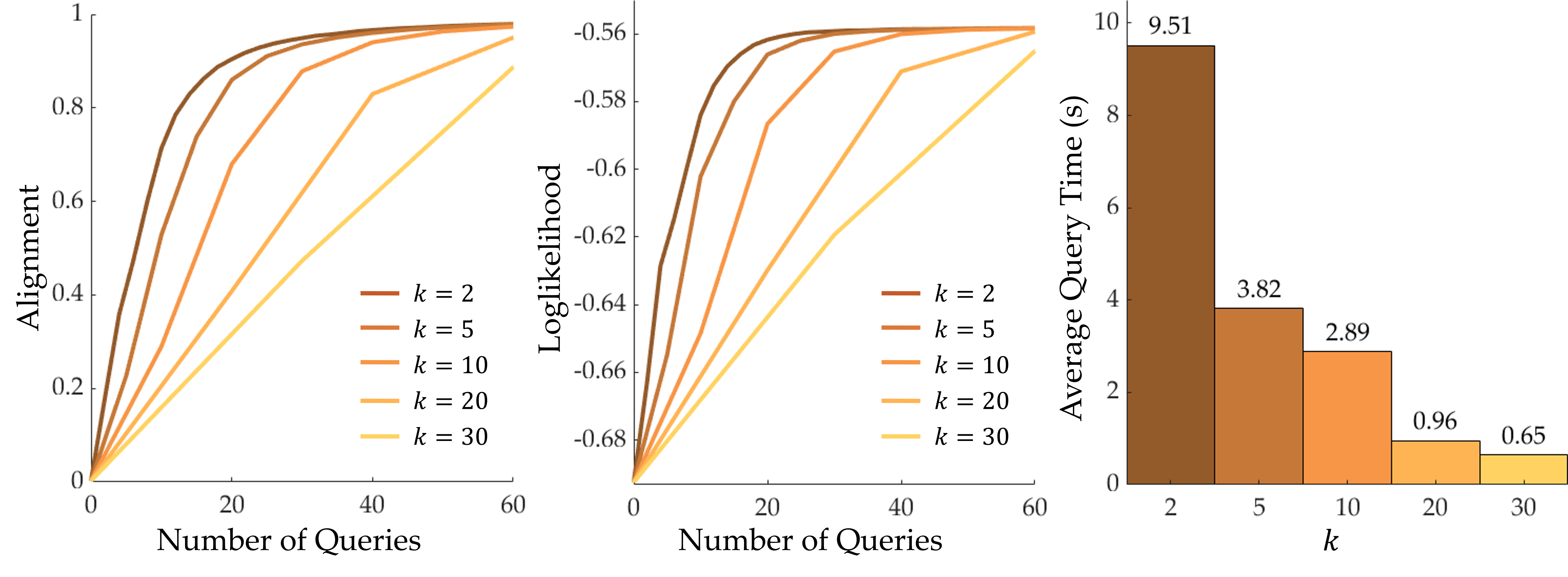}
	\caption{The performance of our DPP-based algorithm with varying $k$ values was averaged over $100$ different runs with \textit{Driver} where $\ww_{\textrm{true}}$ is uniformly randomly generated and $N=20k$. (a) Alignment, (b) loglikelihood, and (c) average query times.}
	\label{fig:batch_size}
\end{figure}

Naturally, non-batch active method would perform better than the DPP-based method if it also selected queries over the discrete set. However, this would require computing mutual information for every query in the set in each and every iteration, which is too slow. Figure~\ref{fig:timing} evaluates the computation time required for querying among our DPP-based method, non-batch active, and random querying by plotting the learning curves during the first $4$ minutes of learning averaged over $100$ seeds. It is clearly visible that batch active learning makes the process much faster than the non-batch active method and random querying. 
Therefore, batch active learning is preferable over other methods as it balances the tradeoff between the number of queries required and the time it takes to compute the queries. This tradeoff can be seen in Figure~\ref{fig:batch_size} where we simulated \textit{Driver} with varying $k$ values. For these simulations, we set $N=20k$ in accordance with other experiments.

\subsection{User Preferences}
In addition to our simulation results using synthetic $\ww_{\textrm{true}}$ vectors, we perform a user study to learn humans' preferences for the  \textit{Driver} and \textit{Tosser} environments. This experiment is mainly designed to show the ability of our framework to learn humans' preferences when human responses may be noisy.

\textbf{Setup.} We recruited $10$ users (ages 18--53, 4 female, 6 male) each of whom responded to $150$ queries generated by successive elimination algorithm under the volume removal acquisition function \cite{sadigh2017active} for each environment (\textit{Driver} or \textit{Tosser}) with batch size $k=10$. None of the users had prior experience with the environments. Users were instructed their goal is to achieve safe and efficient driving in the \textit{Driver} environment. On the other hand, they were told they are free to choose any of the two baskets in the \textit{Tosser} environment as the target (see Figure~\ref{fig:experiment_visuals}~(c)). The goal of this is to demonstrate our batch-active preference based learning algorithms can learn different reward functions that correspond to different behaviors in the environment.

\textbf{Driver Preferences.}
\begin{figure}[t]
	\centering
	\includegraphics[width=0.5\textwidth]{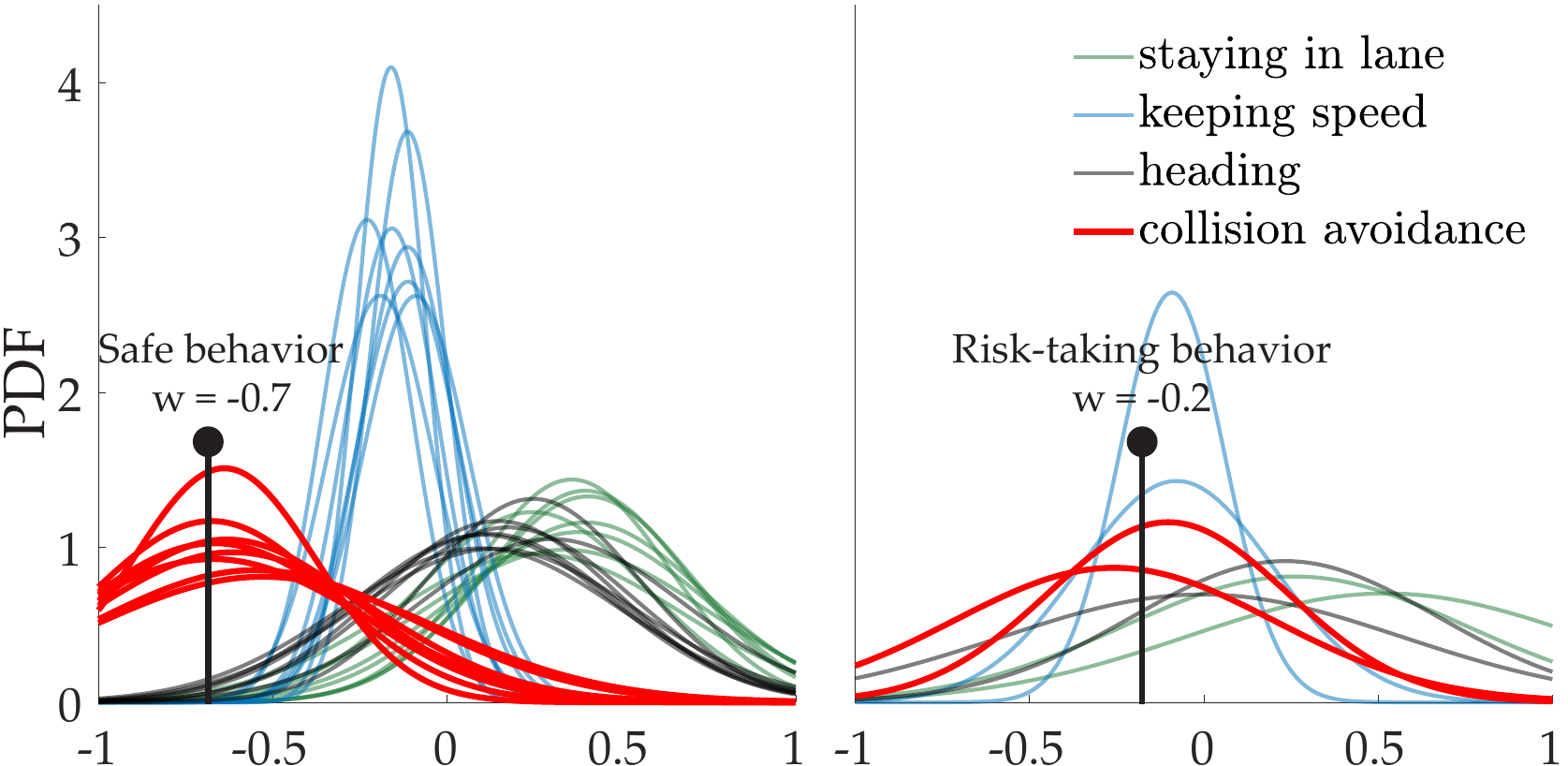}
	\caption{User preferences on \textit{Driver} task are grouped into two sets. While the first set shows the preferences conforming with the natural driving behavior, the second set is comprised of data from two users one of whom preferred collisions with the other car over leaving the road and the other regarded some collisions as near-misses and thought they can be acceptable in order to keep speed. It can be seen that the uncertainty in their learned preferences is higher.}
	\label{fig:user_study_driver}
\end{figure}
Using successive elimination, we are able to learn humans' driving preferences. Our results show that the preferences of users are very close to each other as this task mainly models natural driving behavior. This is consistent with results shown by \cite{sadigh2017active}, where non-batch techniques are used. 
We noticed a few differences between the driving behavior as shown in Figure~\ref{fig:user_study_driver}. This figure shows the distribution of the weights for the four features after $150$ queries. Two of the users (plot on the right) seem to have slightly different preferences about collision avoidance, which can correspond to more aggressive driving behavior.
We observed that $70$ queries were enough for qualitatively converging to safe and sensible driving in the defined scenario. The optimized driving with different number of queries can be watched on \url{https://youtu.be/MaswyWRep5g}. 

\begin{figure*}[t]
	\centering
	\includegraphics[width=\textwidth]{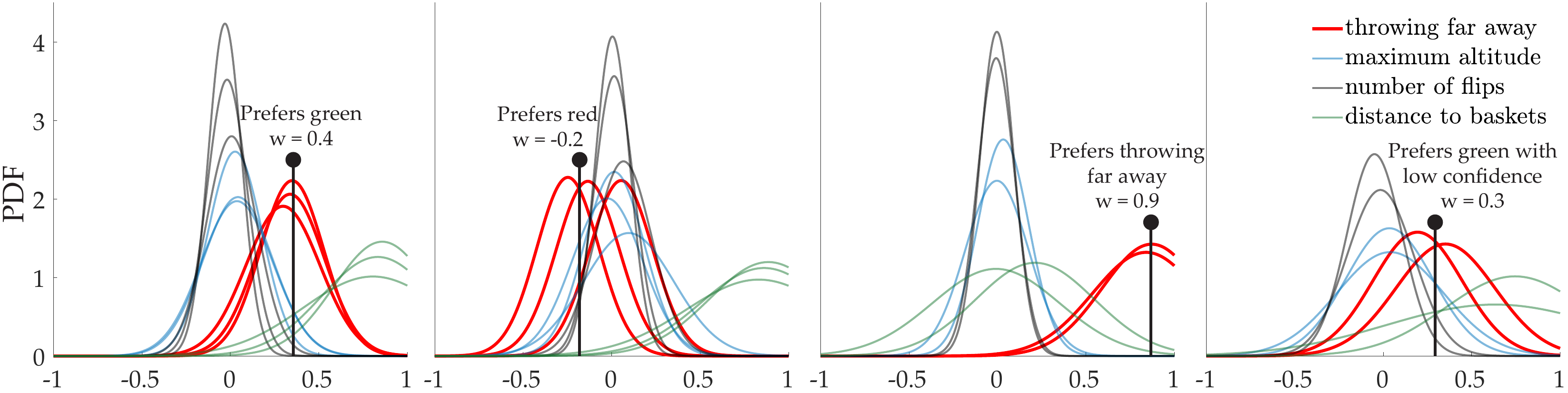}
	\caption{User preferences on \textit{Tosser} task are grouped into four sets. The first set shows the preferences of people who aimed at throwing the ball into the green basket (the distant one) but accepted throwing into the other basket is better than not throwing into any baskets. The second set is comprised of data from three users who preferred the red basket (the closer one). In the third group, the users preferred the green basket over the red one, but also accepted throwing far away is better than throwing into the red basket, because it is an attempt for the green basket. Lastly, the fourth group is similar to the first group; however the confidence over preferences is much less, because the users were not sure about how to compare the cases where the ball was dropped between the baskets in one of the trajectories.}
	\label{fig:user_study_tosser}
\end{figure*}

\textbf{Tosser Preferences.} Similarly, we use successive elimination to learn humans' preferences on the tosser task. Figure~\ref{fig:user_study_tosser} shows we learn interesting tossing preferences, varying between the users based on their target.
For demonstration purposes, we optimize the control inputs with respect to the preferences of two of the users, one of whom prefers the green basket while the other prefers the red one. The evolution of the learning can be watched on \url{https://youtu.be/cQ7vvUg9rU4}. We note that 100 queries were enough to qualitatively see reasonable convergence in this task.


\section{Discussion}

	\quad\textbf{Summary.} In this work, we proposed several end-to-end methods to efficiently learn humans' preferences on dynamical systems. Compared to the previous studies, our method requires only a small number of queries which are generated in a reasonable amount of time. We demonstrated the performance of our algorithms in simulation.
	
	\textbf{Limitations.} 
	In our experiments,	we sampled from the trajectory space in advance for batch active learning methods to generate the dataset of queries $\mathcal{K}$, while we still employed the optimization formulation for the non-batch active version as it was originally proposed by \cite{sadigh2017active}. It can be argued that this creates a bias on the computation times. However, there are two points that make batch active techniques computationally more efficient than the non-batch version. First, they require computing the mutual information values only once in every $k$ queries, whereas the non-batch method requires it for every query. Second, even if we used a pre-generated query dataset for the non-batch active learning method, it would be still inefficient due to adaptive Metropolis algorithm. Furthermore, it can be inferred from Figure~\ref{fig:batch_size} that non-batch active learning with sampling the control space would take a significantly longer running time compared to batch active versions. We also note that query space discretization could reduce the performance of non-batch active learning.


	\textbf{Future directions.} In this study, we used a fixed batch-size. However, we know that the first queries are more informative than the following queries. Therefore, one could start with smaller batch sizes and increase over time. This would both make the first queries more informative and the following queries computationally faster. Hence, further research is warranted to optimize varying batch sizes.
	
	Lastly, we used handcrafted feature transformations, $\phi$'s, in this study.
	In the future we plan to learn those transformations, as in \cite{katz2021preference}, along with the preferences by developing batch techniques that use as few preference queries as possible generated in a short amount of time.

\section*{Acknowledgments}
The authors would like to thank Kenneth Wang for fruitful discussions about the DPP-based batch active learning method; and acknowledge ONR, AFOSR YIP, DARPA YFA, NSF awards \#2218760 and \#2125511, and Ford.

\bibliographystyle{ACM-Reference-Format}
\bibliography{refs}

\appendix

\clearpage
\section{Maximum Coordinate Rounding}\label{app:maximum_coordinate_rounding}
We first present an important point about DPPs that is needed for our maximum coordinate rounding algorithm to approximate the mode of a DPP distribution: conditioning a DPP distribution still results in a DPP. That is, $P(X=A\cup B \mid B\subseteq X)$ is distributed according to a DPP with a transformed kernel:
\begin{align*}
L' = \left(\left[(L + I_{\bar B})^{-1}\right]_{\bar B}\right)^{-1} - I
\end{align*}
where $\bar B = \mathcal{X}\setminus B$, $I$ is the identity matrix, and $I_{\bar B}$ is the projection matrix with all zeros except at the diagonal entries $(i,i)$ for $\forall i\in {\bar B}$ where the entry is $1$.

The greedy approach for approximating the mode of a DPP distribution does not provide the state-of-the-art approximation guarantee. \cite{nikolov2015randomized} showed that one can find an $e^k$-approximation to the mode by using a convex relaxation. We present the algorithm of \cite{nikolov2015randomized} stated in an equivalent form: Formally, consider the generating polynomial associated to the DPP:
\begin{align*}
    g(v_1,\dots,v_K)=\sum_{A:\abs{A}=k}\det(L_A)\prod_{i\in A}v_i\:.
\end{align*}
Finding the mode is equivalent to maximizing $g(v_1,\dots,v_K)$ over \emph{nonnegative integers} $v_1,\dots,v_K$ satisfying the constraint $v_1+\dots+v_K=k$. We get a relaxation by replacing integers with nonnegative reals, and using the insight that $\log(g)$ is a concave function which can be maximized efficiently:
\begin{align*}
\max\left\{\log g(v_1,\dots,v_K)\;\vert\; v_1+\dots+v_K=k \right\}\:.
\end{align*}
If $v_1^*,\dots, v_K^*$ is the maximizer, one can then choose a set $A$ of size $k$ with $P(A)\propto \prod_{i\in A} v_i^*$. Then $\mathbb{E}\left[\det L_A\right]$ will be an $e^k$-approximation to the mode. Although this approximation holds in expectation, the probability that the sampled $A$ is an $e^k$-approximation can be exponentially small. To resolve this, \cite{nikolov2015randomized} resorted to the method of conditional expectations, each time deciding whether to include an element in the set $A$ or not.

The main drawback of this method is its computational cost. In particular, the running time of the methods that compute $g$ scale as a super-linear polynomial in $K$, which is problematic for the typical use cases where $K$ is large. Computing $g$ and $\nabla g$ is needed for solving the relaxation as well as running the method of conditional expectations.

We instead propose a new algorithm that avoids the method of conditional expectations. We also propose a heuristic method to find the maximizers $v_1^*,\dots,v_K^*$ by stochastic mirror descent, where each stochastic gradient computation requires sampling from a DPP (see Appendix~\ref{app:stochastic_mirror_descent}). Approximate sampling from DPPs can be done in time $O(K\cdot k^2 \log k)$, scaling linearly with $K$ \cite{hermon2019modified}. Our algorithm is:
\begin{enumerate}[nosep]
	\item Find the nonnegative real maximizers $v_1^*,\!\dots,v_K^*$ of $\log g(v_1,\!\dots,v_K)$ subject to $v_1+\dots+v_K\!=\!k$.
	\item Let $v_i^*$ be the maximum among $v_1^*,\dots,v_K^*$. Put $i$ in $A$, and recursively find $k-1$ extra elements to put in $A$, working with the conditioned DPP.
\end{enumerate}
\begin{theorem}
	The above algorithm finds an $e^k$-approximation of the mode.
	\begin{proof}
	We prove this by induction on $k$. We simply prove that each time we select an element and put it in $A$, we only lose a factor of at most $e$. Note that the first-order optimality condition of $v_1^*,\dots,v_K^*$ means that
	\begin{align*}
	    \nabla \log g(v_1^*,\dots,v_K^*)=c\mathbf{1}-\sum_{j:v_j^*=0}c_je_j\:,
	\end{align*}
	for some $c$ and collection of $c_j\geq 0$. Here $\mathbf{1}$ is the all-ones vector and $e_1,\dots,e_K$ are the standard basis vectors. By complementary slackness, we have $c_jv_j^*=0$ for all $j$. Since $v_i^*>0$, it must be that $c_i=0$, and it follows that $c=\norm{\nabla \log g(v_1^*,\dots,v_K^*)}_\infty=\partial_i \log g(v_1^*,\dots,v_K^*)$.
	Note that $g$ is a $k$-homogeneous polynomial and it follows that $\dotp{\nabla g(v), v}=kg(v)$. Applying the inequality $\dotp{\nabla g, v}\leq \norm{\nabla g}_\infty\cdot \norm{v}_1$, we get
	\begin{align*}
	    kg(v_1^*,\dots,v_K^*)\leq \norm{\nabla g(v_1^*,\dots,v_K^*)}_\infty\cdot \norm{v^*}_1\:,
	\end{align*}
	Noting that $\norm{v^*}_1=k$ and $\norm{\nabla g(v_1^*,\dots,v_K^*)}_\infty=\partial_i g(v_1^*,\dots,v_K^*)$, we get
	\begin{align*}
	    \partial_i g(v_1^*,\dots,v_K^*)\geq g(v_1^*,\dots,v_K^*)\:.
	\end{align*}
	But note that $\partial_i g$ is exactly the generating polynomial for the conditioned DPP (where we condition on $i\in A$). So it is enough to show that $\max \partial_i g(u_1,\dots,u_K)$ over $u_1+\dots+u_K=k-1$ is at least $1/e$ times the above amount. To do this we simply let $u_j^*=(k-1)v_j^*/(k-v_i^*)$ for $j\neq i$ and we set $u_i^*=0$. Since $\partial_i g$ is $(k-1)$-homogeneous we get
	\begin{align*}
	\partial_i g(u_1^*,\dots,u_K^*) = \left(\frac{k-1}{k-v_j}\right)^{k-1}\partial_i g(v_1^*,\dots,v_K^*) \geq\left(\frac{k-1}{k}\right)^{k-1}g(v_1^*,\dots,v_K^*)\:.
	\end{align*}
	We conclude by noting that $((k-1)/k)^{k-1}\geq 1/e$.
    \end{proof}
\end{theorem}

\section{Stochastic Mirror Descent Algorithm}\label{app:stochastic_mirror_descent}

In this section we propose a stochastic mirror descent algorithm to optimize the following convex program over nonnegative reals
\[ \max\left\{\log g(v_1,\dots,v_N)\;\vert\; v_1+\dots+v_N=k \right\}, \]
where $g$ is the generating polynomial associated to a $k$-DPP, i.e.,
\[ g(v_1,\dots,v_K)=\sum_{A:\abs{A}=k}\det(L_A)\prod_{i\in A} v_i. \]

Our proposed algorithm is repetitions of the following iteration:
\begin{enumerate}[nosep]
	\item Sample a set $A$ with $P(A)\propto \prod_{i\in A} v_i \det(L_A)$.
	\item Let $u\leftarrow v+\eta \mathbf{1}_A$, where $\mathbf{1}_A$ is the indicator of $A$.
	\item Let $v\leftarrow ku/(\sum_i u_i)$.
\end{enumerate}

Note that the sampling in step 1 can be done by Markov chain Monte Carlo (MCMC) methods, since we are sampling $A$ according to a DPP. Careful implementations of the latest MCMC methods (e.g. \cite{hermon2019modified}) run in time $O(K\cdot k^2\log k)$ time. The parameter $\eta$ is the step size and can be adjusted.

Now we provide the intuition behind this iterative procedure. First, let us compute $\nabla \log g$. We have
\[ \frac{\partial_i g(v_1,\dots,v_K)}{g(v_1,\dots,v_K)}=\frac{1}{v_i}\frac{\sum_{A:i\in A}\det(L_A)\prod_{j\in A} v_j}{\sum_{A}\det(L_A)\prod_{j\in A}v_j}, \]
but this is equal to $P(i\in A)/v_i$. Therefore $\nabla \log g=\diag(v)^{-1}p$, where $p$ is the vector of marginal probabilities, i.e. $p_i=P(i\in A)$. Note however that $\mathbb{E}[\mathbf{1}_A]=p$. So this suggests that we can use $\diag(v)^{-1}\mathbf{1}_A$ as a stochastic gradient.

Numerically we found $\diag(v)^{-1}\mathbf{1}_A$ to be unstable. This is not surprising as $v$ can have small entries, resulting in a blow up of this vector. Instead we use a stochastic mirror descent algorithm, where we choose a convex function $h$ and modify our stochastic gradient vector by multiplying $(\nabla^2 h)^{-1}$ on the left.

We found the choice of $h(v_1,\dots,v_K)=\sum_i v_i \log v_i$ to be reasonable. Accordingly, we have $\nabla^2 h =\diag(v)^{-1}$, and therefore
\[ (\nabla^2 h)^{-1}\diag(v)^{-1}\mathbf{1}_A=\mathbf{1}_A. \]
Finally, note that step 3 of our algorithm is simply a projection back to the feasible set of our constraints (according to the Bregman divergence imposed by $h$).

\section{Choice of Stochastic Gradient Vector}

Note that the vector $\mathbf{1}_A$ in step 2 of the algorithm can be replaced by any other random vector $y$, as long as the expectation is preserved. One can extract such vectors $y$ from implementations of MCMC methods \cite{hermon2019modified,anari2019log}. The MCMC methods that aim to sample a set $A$ with probability proportional to $\prod_{i\in A} v_i \det(L_A)$ work as follows: starting with a set $A$, one drops an element $i\in A$ chosen uniformly at random, and adds an element $j$ back with probability proportional to $\det L_{A-i+j}$, in order to complete one step of the Markov chain. We can implement the same Markov chain, and let $y_j$ be $k$ times the probability of transitioning from $A-i$ to $A-i+j$ in this chain. It is easy to see that if the chain has mixed and $A$ is sampled from the stationary distribution
\begin{align*}
    \mathbb{E}[y]=\mathbb{E}[\mathbf{1}_A]\:.
\end{align*}
We found this choice of $y$ to have less variance than $\mathbf{1}_A$ in practice.

\section{Empirical Comparison of Maximum Coordinate Rounding with Greedy Approach}
Here we provide an empirical comparison between the performance of the greedy approach to approximating the mode of a DPP distribution versus our maximum coordinate rounding algorithm.

\begin{figure}
	\centering
	\includegraphics[width=0.7\textwidth]{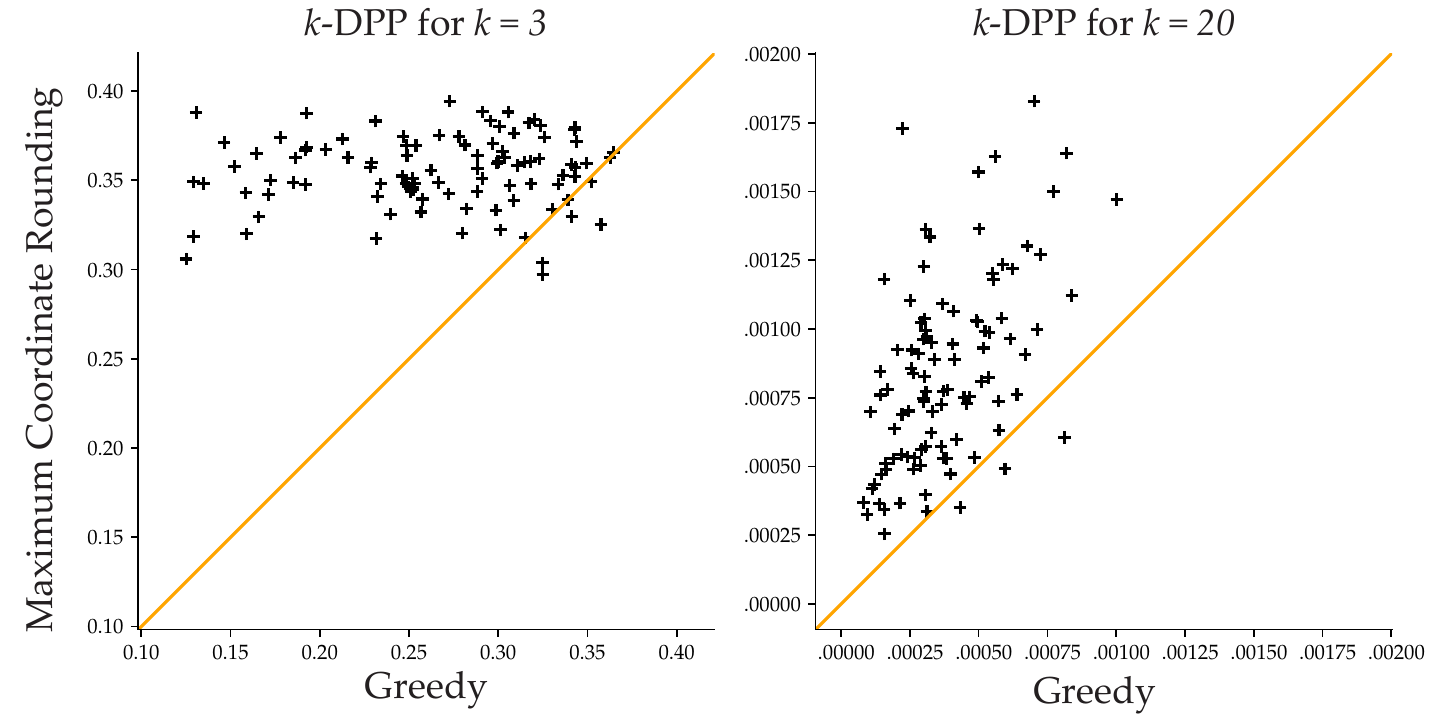}
    \vspace{-10px}
	\caption{Comparison of the greedy algorithm and the maximum coordinate rounding algorithm. In 93\% of the $k\!=\!3$ cases, and in 97\% of the $k\!=\!20$ cases, our method returns a better or equal solution.}
    \vspace{-15px}
	\label{fig:convex-comparison}
\end{figure}

We used two sets of experiments where $q_i=1$ for all $i$ in the query set. In the first, we generated $200$ random queries inside $[0,1]^2$ (meaning their feature difference vectors, $\psi$'s, are inside $[0,1]^2$), set $\sigma=1$ and attempted to find the mode of the $k$-DPP for $k=3$. In the second, we generated $200$ random queries inside $[0,1]^2$, set $\sigma=0.2$ and attempted to find the mode of the $k$-DPP for $k=20$. We ran each experiment $100$ times (each time generating a new set of random queries).

The results can be seen in Fig.~\ref{fig:convex-comparison}. We plotted $\det(L_A)$ vs.\ $\det(L_B)$, where $A$ is the set returned by the greedy approach, and $B$ is the set returned by our maximum coordinate rounding algorithm.

\end{document}